\documentclass[10pt,journal,compsoc]{IEEEtran}
\ifCLASSOPTIONcompsoc
\usepackage[nocompress]{cite}
\else
\usepackage{cite}
\fi
\ifCLASSINFOpdf
\else
\fi
\ifCLASSINFOpdf
\usepackage[pdftex]{graphicx}
\graphicspath{{./images}}
\DeclareGraphicsExtensions{.pdf,.jpeg,.jpg,.png,.PDF,.JPEG,.PNG}

\usepackage{ifpdf}
\usepackage{cite}
\usepackage{xcolor}
\usepackage{amsmath}
\usepackage{amssymb}
\usepackage{amsthm}
\usepackage{amsfonts}
\usepackage{algorithm}
\usepackage{algorithmic}
\usepackage{multirow}
\usepackage{array}
\usepackage{booktabs}
\usepackage{hyperref}
\usepackage{bm}
\ifCLASSOPTIONcompsoc
\usepackage[caption=false,font=normalsize,labelfont=sf,textfont=sf]{subfig}
\else
\usepackage[caption=false,font=footnotesize]{subfig}
\fi

\usepackage{stfloats}

\usepackage{url}

\usepackage[switch]{lineno}

\DeclareMathOperator*{\argmin}{\arg\min}

\def\L{{\cal L}}

\def\i{{\mathrm{p}}}
\def\j{{\mathrm{q}}}

\def \TODO [#1]{\textcolor{red}{TODO: #1}}
\def \NOTE [#1]{\textcolor{blue}{(\textit{#1})}}
\def \comment [#1]{}
\def \etal {\emph{et. al.~}}

\def \check {$\checkmark$}
\def \cross {$\times$}

\newtheorem{definition}{Definition}

\begin{document}
%
\title{Implicit Subspace Prior Learning for\\ Dual-Blind Face Restoration}
%
%
%
%

\author{Lingbo~Yang,
	Pan~Wang, Zhanning~Gao, Peiran~Ren, Shanshe~Wang,
	Siwei~Ma,~\IEEEmembership{Member,~IEEE}
	and~Wen~Gao,~\IEEEmembership{Fellow,~IEEE}
	\thanks{Lingbo Yang, Shanshe Wang, Siwei Ma and Wen Gao are from the National Engineering Laboratory of Video Technology (NELVT), Peking University.
		Siwei Ma is the corresponding author.
		Email: \{lingbo, sswang, swma, wgao\}@pku.edu.cn
		
		Zhanning Gao, Pan Wang, Peiran Ren are from the Alibaba Group.
		
}}

%
%

\markboth{Journal of \LaTeX\ Class Files,~Vol.~14, No.~8, August~2020}%
{Shell \MakeLowercase{\textit{et al.}}: Bare Demo of IEEEtran.cls for Computer Society Journals}
%



\IEEEtitleabstractindextext{%
	\begin{abstract}
		Face restoration is an inherently ill-posed problem, where additional prior constraints are typically considered crucial for mitigating such pathology. However, real-world image prior are often hard to simulate with precise mathematical models, which inevitably limits the performance and generalization ability of existing prior-regularized restoration methods. In this paper, we study the problem of face restoration under a more practical ``dual blind'' setting, i.e., without prior assumptions or hand-crafted regularization terms on the degradation profile or image contents.
		To this end, a novel implicit subspace prior learning (ISPL) framework is proposed as a generic solution to dual-blind face restoration, with two key elements: 1) an implicit formulation to circumvent the ill-defined restoration mapping and 2) a subspace prior decomposition and fusion mechanism to dynamically handle inputs at varying degradation levels with consistent high-quality restoration results.
		Experimental results demonstrate significant perception-distortion improvement of ISPL against existing state-of-the-art methods for a variety of restoration subtasks, including a 3.69db PSNR and 45.8\% FID gain against ESRGAN, the 2018 NTIRE SR challenge winner. Overall, we prove that it is possible to capture and utilize prior knowledge without explicitly formulating it, which will help inspire new research paradigms towards low-level vision tasks.
	\end{abstract}
	
}

\maketitle

\IEEEdisplaynontitleabstractindextext

\ifCLASSOPTIONpeerreview
\begin{center} \bfseries EDICS Category: 3-BBND \end{center}
\fi
%
\IEEEpeerreviewmaketitle

\IEEEraisesectionheading{\section{Introduction}\label{sec:introduction}}

\IEEEPARstart{F}{ace} restoration, which aims to generate high-quality (HQ) face images from low-quality (LQ) observations, is a typical inverse problem in low-level vision. 
Although significant progress has been made under specific prior assumptions~\cite{srcnn}\cite{sr_res_attn}\cite{esrgan}\cite{DeblurGAN}\cite{RIDNet}\cite{waveletcnn_conf}\cite{progressive-face-sr}\cite{SPANet}\cite{DA_dehazing}, it still remains an open problem to design robust and versatile face restoration methods under a more generalized problem setting.

Specifically, existing face restoration methods often assume the LQ input $y$ is generated from an underlying clean image $x$ with a fixed degradation model $\mathcal{D}$, i.e., $y=\mathcal{D}(x)$. For most real-world scenarios, the degradation $\mathcal{D}$ is usually irreversible, making face restoration an inherently ill-posed problem where for any given $y$, there could exist multiple predictions $\hat{x}$ all satisfying the condition $\mathcal{D}(\hat{x})=y$. To reduce such ambiguity, existing works often introduce additional prior regularization on either the degradation $\mathcal{D}$ or image content distribution $P_X$, leading to the following \emph{prior-regularized restoration problem}:

\begin{equation}
	\tilde{x} = \argmin_{x} \Delta(\mathcal{D}(x), y) + \lambda_D\mathcal{R}_D(\mathcal{D}) + \lambda_c \mathcal{R}_c(x;P_X)
	\label{eqn:lagrange_formulation}
\end{equation}
where $\Delta(.,.)$ is a distortion measure such as MSE or L1 loss, and $\mathcal{R}_D$ and $\mathcal{R}_c$ are regularization terms on degradation and content prior, respectively. 

\begin{figure*}[!t]
	\centering
	\includegraphics[width=\linewidth]{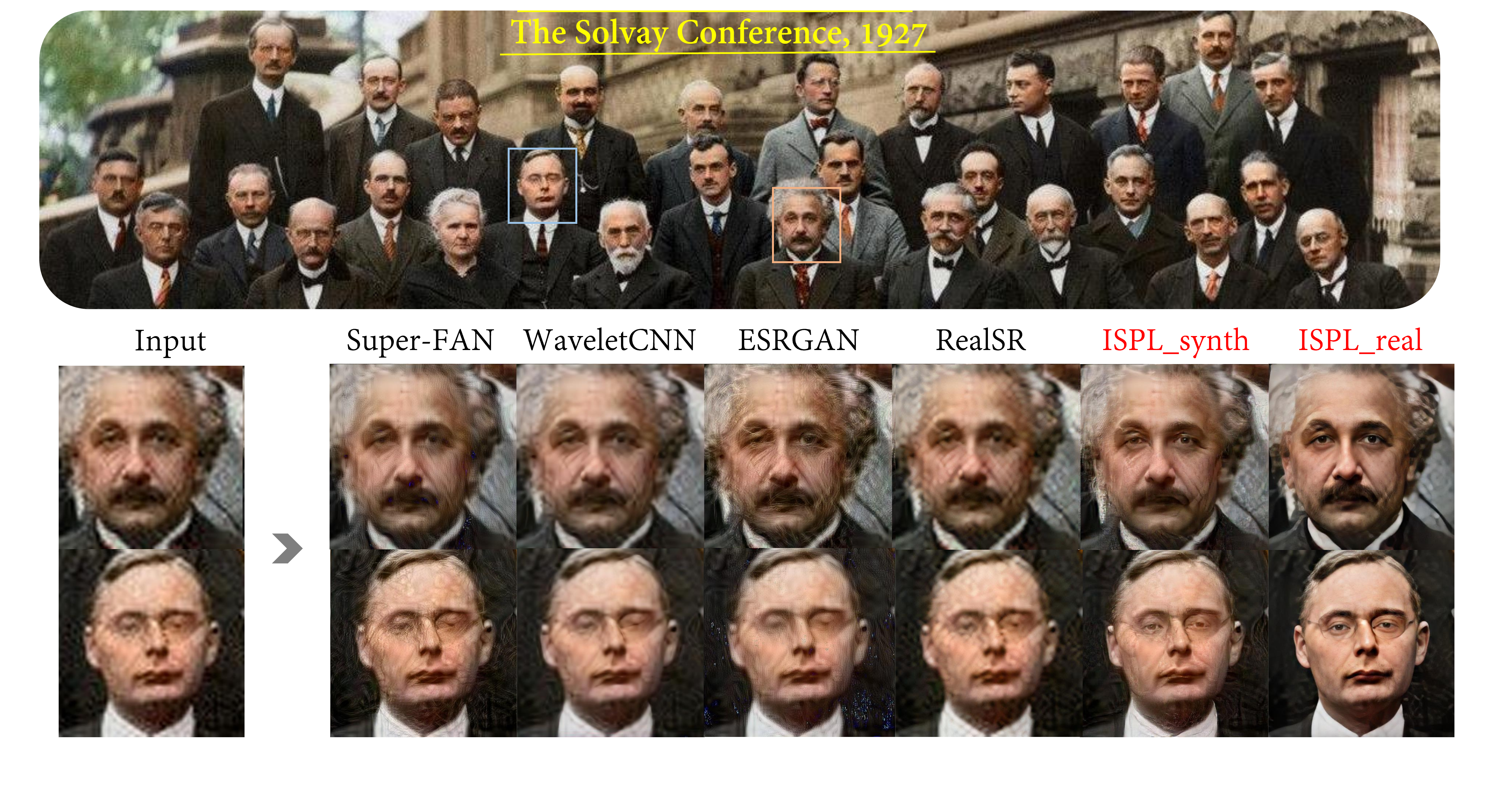}
	\caption{\textbf{Comparison of face restoration results on real-world faces.} Existing methods with explicit degradation prior assumption, including the NTIRE2020 winner RealSR~\cite{real-worldSR}, all fail to cope with such complex degradation, leaving noise residuals and additional artifacts in the restored image. In contrast, our proposed implicit subspace prior learning (ISPL) algorithm produces superior results even when trained on the same synthetic dataset (ISPL synth), and is further empowered  after training with our new FaceRenov dataset (ISPL real). \emph{(Best to view on the computer screen for your convenience to zoom in and compare the quality of facial details. Ditto for other figures.)}}
	\label{fig:teasor}
\end{figure*}

For degradation prior modeling, existing restoration methods mostly target at primitive degradation types, such as downsampling, noise, blur, compression artifacts, and so on. Due to the complexity in real-world scenes, it is often difficult to simulate real-world degradation prior precisely with a simple combination of these primitive functions. In consequence, models trained upon synthetic data often exhibit poor generalization towards real-world samples. Fig.~\ref{fig:teasor} showcases several state-of-the-art restoration methods, including RealSR~\cite{real-worldSR}, the winner of 2020 NTIRE super resolution challenge~\cite{NTIRE2020}, over the famous group photograph of the 5th Solvay conference. Clearly, all methods trained on synthetic data tend to produce noticeable artifacts, despite that ours are considerably milder than other baselines. In contrast, after training on a more realistic dataset (as detailed in Sec.~\ref{sec:facerenov}), our method shows better adaptability over real-world samples, with more prominent facial landmarks and less artifacts. This empirically suggests the significant domain gap between real-world and simulated degradation, as well as its non-negligible impact over degradation-specific (a.k.a. ``non-blind'') restoration methods.

\begin{figure*}[!t]
	\centering
	\includegraphics[width=\linewidth]{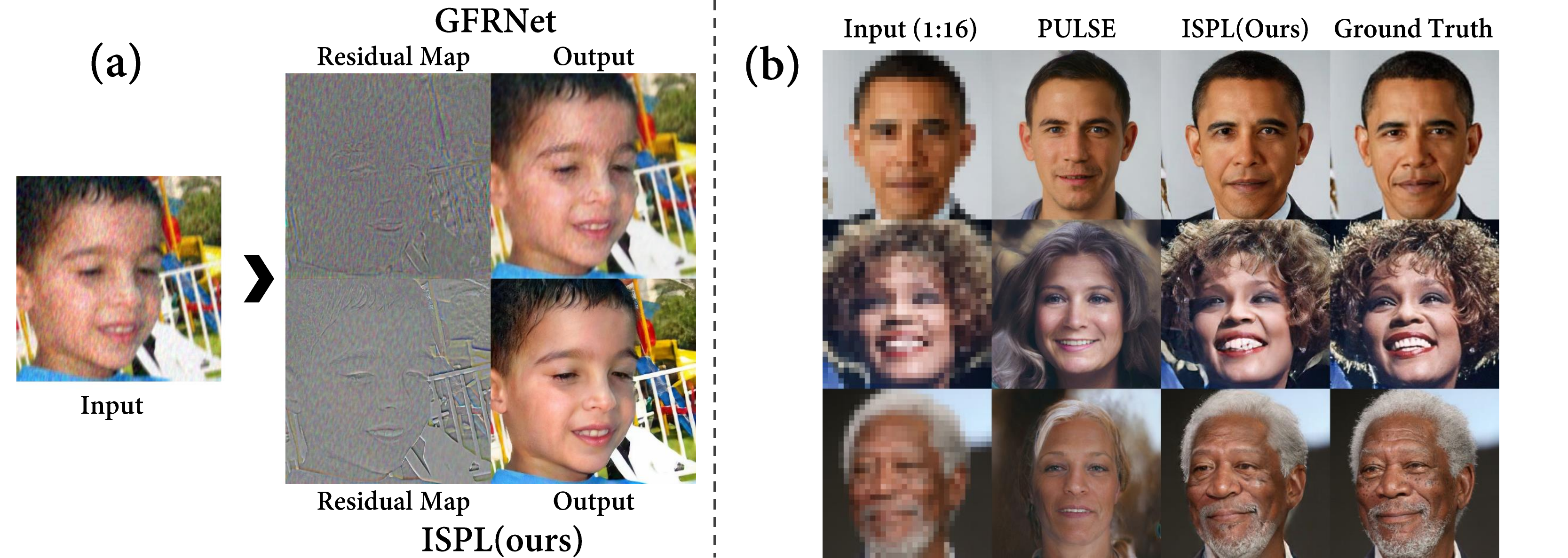}
	\caption{Two typical examples demonstrating the limitation of explicit prior learning schemes. (a) GFRNet~\cite{BlindFR-ECCV2018} with explicit facial prior regularization (landmark loss) restricts the ability to enhance background objects. (b) PULSE~\cite{PULSE} with a biased face generator trained on Caucasian faces failed to generalize on colored celebrities, leading to the observed ``racial bias''. In contrast, our ISPL resolves the above issues outstandingly.}
	\label{fig:weakness}
\end{figure*}

When it comes to content prior modeling, existing methods either enforce semantic or structural consistency between the restored image and HQ reference via feature matching loss~\cite{ComponentSP}\cite{face_hallu_1}\cite{SuperFAN}\cite{FSRNet}, or utilize additional exemplar patches to guide the reconstruction of realistic facial details, such as GFRNet~\cite{BlindFR-ECCV2018} or DFDNet~\cite{BlindFR-ECCV2020}. However, applying a face-oriented regularization often impairs the restoration capability of non-facial objects and background contents, which are usually abundant in real-world images. Fig.~\ref{fig:weakness}(a) compares the residual map generated from GFRNet~\cite{BlindFR-ECCV2018} and our method, where GFRNet is much less effective in renovating background contents, thus producing inferior restoration results.
Another line of works attempt to restrict the solution space onto a pretrained latent manifold of HQ faces, and approach restoration via latent exploration~\cite{Image2StyleGAN}\cite{Guanshanyan2020Collaborative}\cite{DeepSEE}\cite{PULSE}. Despite the improved realism, the accuracy of latent embedding is often susceptible to degradation artifacts, which could result in observable content discrepancy in restored faces. Fig.~\ref{fig:weakness}(b) demonstrates a typical failure case of ``racial bias''~\cite{PULSE_racism} where several colored celebrities are falsely converted to Caucasian, including Barack Obama, the first African-American president. In contrast, our method well preserves the ethnicity of input faces, albeit trained upon the same biased dataset. 

The above observation suggests, despite our common sense, that explicit prior regularization is not always beneficial for face restoration. Instead, the complexity in real-world scenarios often leads to partial or inaccurate prior modeling~\cite{SimultaneousFnRLearning}, which would degrade the restoration performance and generalizability. To address this issue, we propose to investigate a novel ``dual-blind'' setting of the face restoration problem, which lifts the prior constraints on either the degradation process or underlying image contents. In this way, the restoration model can more flexibly adapt to datasets with complex, unidentified degradation profile, where the relationship between LQ and HQ images can be captured in an implicit fashion. Moreover, by lifting the prior assumption, dual-blind restoration will lead to a unified solution to arbitrary degradation types that would traditionally require meticulous network and loss design for each individual case.
This leads to a natural question: \emph{Is it possible (and how) to capture the intrinsic prior knowledge from observed data and utilize it for restoration, without explicitly modeling it?}

In this paper, we propose a novel implicit subspace prior learning (ISPL) framework to approach dual-blind face restoration, with two major distinctions from previous restoration methods: 1) Instead of assuming an explicit degradation function between LQ and HQ domain, we establish an implicit correspondence between both domains via a mutual embedding space, thus avoid solving the pathological inverse problem directly. In this way, our ISPL captures and utilizes the prior knowledge for any restoration subtask in a unified, data-driven manner, without extra feature or loss engineering.
2) Instead of assuming a fixed embedding space, we design a flexible subspace structure to handle testing samples at varying degradation levels with consistent restoration performance. Furthermore, we propose a dynamic prior fusion module to convert the discrete architecture search into a continuous learning problem, allowing dynamic perception-distortion tradeoff~\cite{Tradeoff} at testing stage without fine-tuning or iterative optimization as required by most traditional works~\cite{Tradeoff}\cite{PULSE}.

The proposed ISPL covers a wide range of powerful generative models, including U-net, StyleGAN~\cite{stylegan_ffhq}, and multi-code GAN prior~\cite{mganprior}, while at the same time allows flexible ensemble of different models. 
As a result, ISPL surpasses existing restoration methods by a significant margin over five well-studied restoration subtasks and our new dual-blind face restoration. Moreover, we develop a new benchmark dataset, FaceRenov, to quantify the generalization capability of ISPL towards real-world scenarios, which also provides further insight to the challenges and possible future directions for real-world face restoration. Overall, our major contributions are threefold:

\begin{itemize}
	\item We investigate a novel ``dual-blind'' setting for face restoration, which aims for a unified solution to resolve arbitrary degradation in a data-driven approach without explicit prior regularization. To the best of our knowledge, we are the first ones to study, and empirically prove the superiority of implicit formulation of face restoration against explicit ones, which could inspire a new research methodology for low-level vision tasks.
	
	\item We propose a novel implicit subspace prior learning (ISPL) framework to  utilize prior knowledge for face restoration without explicitly formulating it, which enjoys superior adaptability to complex degradation against prior-regularized methods, and achieves dynamic perception-distortion tradeoff at testing stage.
	
	\item We conduct extensive experiments over five restoration subtasks, verifying the superior perception-distortion performance and better generalizability of implicit prior learning over existing restoration methods.
\end{itemize}

This manuscript extends upon our previous work~\cite{yang2020hifacegan} in three aspects: 
First of all, we establish the theoretical foundation for dual-blind face restoration, and incorporate our previous model (dubbed as ``HiFaceGAN'') under the more general ISPL framework as a specific variant. Secondly, we propose an effective dynamic prior fusion module to convert the discrete subspace architecture search problem into a continuous one, allowing flexible perception-distortion tradeoff against varying testing conditions without fine-tuning the restoration model. 
Finally, we propose a new benchmark dataset, FaceRenov, to analyze of the domain gap between real-world and simulated problem setting, as well as the generalization capability of different restoration models. Both qualitative and quantitative experiments are conducted on this dataset to further justify the proposed implicit prior formulation for restoration tasks.
Although primarily designed for face restoration, ISPL is equally effective in more complex cases, such as natural scene images and video clips, which could inspire new research paradigms for other low-level vision tasks. Code is available at \url{https://github.com/Lotayou/Face-Renovation}.

\section{Related Works}


Degradation and content prior are two crucial factors in face restoration, where a practical system is expected to cope with varying testing conditions. As such, extensive efforts have been put into accurate and flexible design of prior models. In this section, we review existing face restoration methods based on their prior modeling schemes, both over the degradation process or underlying image contents. Meanwhile, recent generative-based restoration methods are also reviewed under the prior-learning perspective.

\subsection{Degradation Prior Modeling}

Existing restoration methods typically aim at specific degradation types, leading to a variety of restoration subtasks, including super resolution~\cite{sr_review}, denoising~\cite{denoising_survey}, deblurring~\cite{deblurring_survey}, to name a few. For these subtasks, the training data is typically synthesized via a fixed degradation model composed of one or more basic mathematical operations, including convolution, downsampling, additive noise and JPEG compression~\cite{BlindFR-ECCV2018}:

\begin{equation}
	y = \mathcal{D}(x; \mathbf{k}, s, \sigma, q) = \mathcal{J}_q((x \circledast \mathbf{k})\downarrow_s + \mathbf{n}_\sigma)
	\label{eqn:deg}
\end{equation}

where $\mathbf{k}, s, \sigma, q$ are degradation parameters that are either fixed in advance or to be estimated from training data. Traditionally, the restoration model and loss design heavily depends on the prior assumption, leading to limited transferring ability across different settings, even within the same degradation category. For instance, a super resolution model pretrained under 4x scaling can also fail at 8x or 16x cases, Fig.~\ref{fig:pd_tradeoff}.
To address this issue, blind image restoration~\cite{BIR_original} employs a deep neural network with learnable parameters to achieve a more flexible and accurate prior modeling, such as blind kernel estimation with adaptive convolution~\cite{Blind-SR-kernel}\cite{Deep-plugnplay} or iterative kernel correction\cite{Kernel-modeling-SR} from observed image pairs. However, despite the relative ease in model and loss engineering, existing blind restoration methods still relies on precise degradation modeling, and are quite sensitive to the perturbation of testing conditions~\cite{BlindIR-wo-prior}. Adrian \etal~\cite{learn_degrade} learns the degradation and restoration model simultaneously from unpaired training data, yet the efficacy is only verified for low resolution faces ($64\times64$). Ren \etal~\cite{SimultaneousFnRLearning} propose simultaneous fidelity and regularization learning to compensate for partial or inaccurate degradation prior modeling, which involves a complicated iterative optimization of hand-crafted regularization terms. Different from existing methods, we propose an implicit, ``dual-blind'' approach to face restoration without relying on explicit prior models or regularization terms, leading to superior performance over each individual restoration subtask and better generalizability across different testing conditions.

Additionally, a recent work by Elron \etal~\cite{BlindIR-wo-prior} also explores the blind image restoration problem without prior knowledge. Their solution is based on a self-normalization side-chain (SCNC) that automatically detects relevant imaging parameters from observable data. While sharing the same ``prior-free'' assumption, our work focus on the collaborative modeling of degradation and content prior instead of just the degradation process alone. Moreover, at testing stage, SCNC requires additional tuning to cope with different testing conditions, whereas our ISPL framework supports dynamic perception-distortion tradeoff, and can handle a reasonable degradation variation without fine-tuning model parameters.


\subsection{Content Prior Modeling}
Human face carries strong semantic and structural prior that is usually considered beneficial for improving the restoration performance. A common prior utilization scheme is through the feature matching loss between the restored face and the corresponding HQ supervision:

\begin{equation}
	\mathcal{R}_c(x) = \|\phi(\tilde{x}) - \phi(x)\|\label{eqn:content_prior}
\end{equation}

where $\phi(.)$ is a corresponding function to extract informative features, such as component semantic prior~\cite{ComponentSP}, identity~\cite{face_hallu_1}, facial landmarks~\cite{SuperFAN}\cite{FSRNet} or heatmaps~\cite{faceSR_ECCV2018}. Besides feature matching, other works also regulate the restoration model with objectives of subsequent vision tasks, such as semantic segmentation~\cite{whenID_meets_high} and image recognition~\cite{Joint_BIR_recog}. Yet such top-level regularization does not necessarily guarantee the quality of low-level visual details in restored outputs. Alternatively, exemplar-based restoration methods~\cite{BlindFR-ECCV2018}\cite{EnhancedBlindFR-CVPR2020} aim to use additional HQ image(s) of the same person for guided facial detail enhancement, but this is intractable for restoring unidentified faces. Sparse dictionary based methods~\cite{LRFR-sparse}\cite{BlindFR-ECCV2020} aim to reconstruct facial details based on a group of learned dictionaries of exemplar patches, with each dictionary corresponding to a particular facial landmark. However, the performance of exemplar-based methods depends heavily on the dictionary capacity, which is often limited at resource-constrained scenarios, e.g. mobile applications. In comparison, our ISPL learns the characteristics of HQ face images implicitly into the generative restoration model parameters, allows it to directly synthesize realistic facial details directly from partial/inaccurate observations. In this way, ISPL can achieve stunning perceptual quality for severely degraded input with no extra supervision.

\subsection{Deep Generative Restoration models} 
Recent advances in high-resolution face generation~\cite{progressivegan}\cite{stylegan_ffhq} has opened up a new approach to face restoration that explores a pretrained manifold of HQ faces to find the best match to the LQ input. To acquire the latent representation, some works aim to learn an encoder that projects the degraded input onto HQ manifold~\cite{Image2StyleGAN}\cite{Image2StyleGAN++}~\etal\cite{Guanshanyan2020Collaborative}, while others perform gradient-based optimization over the latent manifold to minimize the reconstruction error (Eqn.~\eqref{eqn:lagrange_formulation}), including StyleGAN2~\cite{stylegan2} and PULSE~\cite{PULSE}. Generally, the latent embedding method allows more flexible semantic exploration and attribute editing~\cite{DeepSEE}, and can produce plausible faces under severe degradation (\emph{e.g.} 64 times downsampling) beyond the limit of existing reconstruction-based methods. However, a single latent space is often inadequate to capture the diversity of real-world faces, and the corresponding latent vector could be susceptible to input perturbation. To solve this issue, Gu.~\etal\cite{mganprior} use multi-gan codes to achieve disentangled prior learning, and B\"{u}hler~\etal\cite{DeepSEE} further incorporate parsing maps to also allow expression editing with semantic region adaptive normalization~\cite{sean}. Also, invertible neural networks~\cite{SRFlow}\cite{InvertibleIR} are utilized to model the conditional distribution $p_{\tilde{X}|Y}$ instead of a single output $\tilde{x}$, which inherently resolves the pathology of the original ``one-to-many'' problem. 

Compared to the methods above, our ISPL follows the same ``embedding \& generation'' paradigm, yet adopts an implicit, data-driven learning approach to automatically determine the optimal latent decomposition and aggregation rules as opposed to hand-crafted ones. In this way, ISPL can better adapt to complex, unidentified prior conditions, leading to superior restoration performance against existing methods on real-world images.

\begin{figure*}
	\centering
	\includegraphics[width=\linewidth]{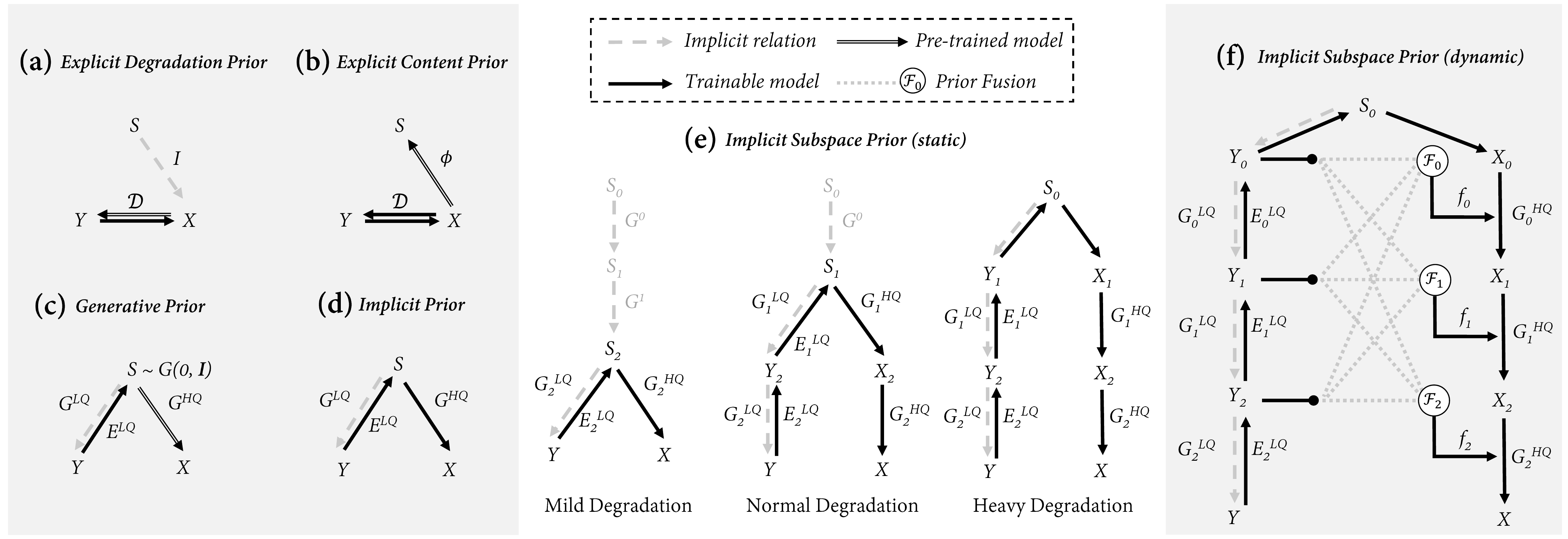}
	\caption{\textbf{Illustration of different prior formulations for face restoration.} Compared to explicit prior models~(a)~(b)~(c), our implicit model (d) connects LQ and HQ domain via a common embedding space S, which can be further decomposed into subspaces (e) and splitted across LQ/HQ domain to deal with varying degradation levels. Finally, (f) shows a unified architecture with dynamic prior fusion for flexible ensemble learning. \emph{Different elements are separated with gray boxes for better visual effects.}}
	\label{fig:framework}
\end{figure*}

\section{Implicit Subspace Prior Learning}
\label{sec:ISPL}

In this section, we develop implicit subspace prior learning (ISPL) as a general framework for our dual-blind face restoration problem, and discuss its relation and distinction between existing methods. Possible implementation options are outlined in this section, with an ablation study conducted in Sec~\ref{sec:ablation} to determine the optimal design from several representative architectures. Furthermore, the subspace decomposition and accumulative restoration results are visualized at each subspace, which reveals the working mechanism of ISPL and justifies the heuristics in model design.

\subsection{An Implicit Formulation for Face Restoration}

Given a low-quality (LQ) face $y$, we aim to reconstruct a high quality (HQ) face $\tilde{x}$ that matches the partial information observable in $y$ without assuming specific degradation prior $P_{Y|X}$ or content prior $P_X$ in advance. To this end, an implicit correspondence is established via a common embedding space $S$ between both LQ and HQ image domain, which are defined with independent generative mappings:

\begin{equation}
	\label{eqn:embed}
	X = G^{HQ}(S);\quad Y = G^{LQ}(S)
\end{equation}

Fig.~\ref{fig:framework} (a)-(d) illustrates the relationship between explicit restoration methods and our implicit formulation. For methods assuming an explicit degradation prior (a), they are equivalent to the trivial case of $S=X, G_{HQ} = I:x\to x$ and $G_{LQ}=\mathcal{D}$. Consequently, the restoration mapping degenerates to the inverse of $\mathcal{D}$, which is inherently ill-defined. For methods with explicit content prior (b), the latent embedding space is determined by a pretrained feature extractor $\phi$,  Eqn.~\eqref{eqn:content_prior}.
For generative restoration methods (c), the high-quality mapping $G_{HQ}$ is usually fixed to a pretrained generative model (for instance, StyleGAN) with a prescribed latent distribution of $S$, such as Gaussian. However, this often limits the representative capacity for complex real-world scenes. In contrast, our implicit prior formulation~(d) approaches face restoration via a \emph{collaborative} learning of semantic embedding $E^{LQ}$ and guided face restoration $G^{HQ}$, which can autonomously capture and utilize the prior knowledge without relying on hand-crafted degradation models or regularization terms as in most explicit methods.

\subsection{Subspace Decomposition}
Generally, the structure of common embedding space $S$ would vary at different degradation levels, where heavier degradation indicates more severe information loss and signal distortion, which implies larger content discrepancy between LQ and HQ domain and consequently, a ``smaller'' embedding space $S$.  
As a general guideline, the scale-space theory~\cite{ScaleSpaceTheory} states that image contents across different scales (i.e. degradation levels) follows a strict hierarchy, with coarse-level scale-space interest points being a subset of the ones at finer scales. It is thus reasonable to decompose $S$ into a set of hierarchical subspaces for handling different cases in a unified manner. Concretely, we can upgrade Eqn.~\ref{eqn:embed} to the following \emph{implicit subspace prior model}:

\begin{definition} An implicit subspace prior (ISP) model with n total layers and k shared ones, denoted as $ISP(k,n)$, are defined as follows:
	\begin{equation}
		S^k = \prod_{i=0}^{k-1}G_i(S^0);\quad X = \prod_{i=k}^{n-1}G_i^{HQ} (S^k);\quad Y =\prod_{i=k}^{n-1}G_i^{LQ}(S^k)
		\label{def:isp}
	\end{equation}	
\end{definition}

Here we slightly abuse the product notation to denote function composition, so that $\prod_{i=0}^{k-1}G_i = G_{k-1} \circ G_{k-2} \circ\cdots\circ G_1 \circ G_0$. Now the original formulation in Eqn.~\ref{eqn:embed} becomes the trivial case of $ISP(0,1)$, where the embedding space $S$ is fixed. 
In contrast, ISP model can cope with varying degradation by adjusting the structure of embedding space via $k$, with smaller $k$ accounting for heavier degradation that incurs larger content discrepancy. Fig.~\ref{fig:framework}(e) visualizes the diagram of $ISP(k,3)$, with $k=2,1,0$ for mild, normal, and heavy degradation, respectively.\footnote{The inner subspaces $S_0, \dots, S_{k-1}$ helps clarify the relationship between different cases, yet are not actually implemented in practice.}
The restoration is then defined via an embedding function $E^{LQ}$ that projects $y$ onto the k-th common subspace $S^k$ at $y_k = E^{HQ}(y)$, from which the restoration result $\tilde{x}$ is generated. Mathematically,

\begin{equation}
	\label{eqn:ISP}
	\tilde{x} = G^{HQ}(E^{LQ}(y)) = \prod_{i=k}^{n-1}G_i^{HQ} \circ \prod_{j=n-1}^{k}E_j^{LQ} (y)
\end{equation}

\textbf{Relationship to Perception-Distortion Tradeoff}
Blau~\etal\cite{Tradeoff} propose a systematic way to explore the perception-distortion (P-D) tradeoff in image restoration by adjusting the balance between training losses, which is often intractable at testing stage. In contrast, our ISP model provides an alternative tradeoff scheme by altering the subspace level $k$, where a smaller $k$ corresponds to more perceptually-inclined restoration strategy that emphasizes less on input features and more on prior-guided detail synthesis. 
Powered by the dynamic prior fusion mechanism, our ISPL can support dynamic P-D tradeoff against varying degradation levels without model tuning. More analysis and empirical evidences will be detailed in Sec~\ref{sec:pd_tradeoff}.

\begin{figure*}
	\centering
	\includegraphics[width=1\linewidth]{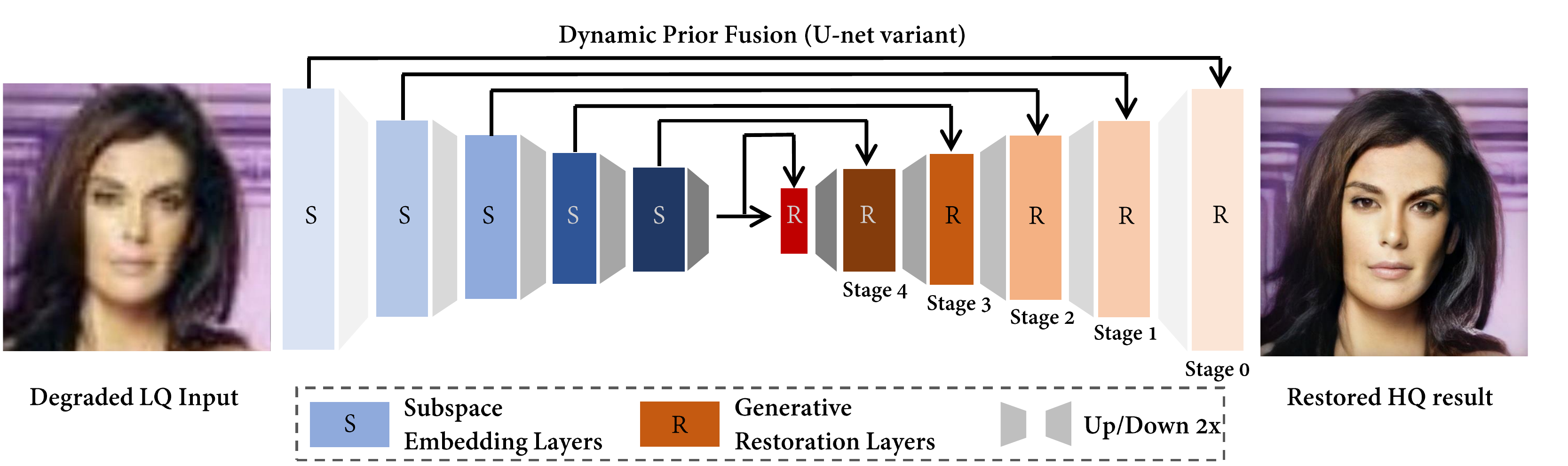}
	\caption{Illustration of our benchmark solution, ISPL-Unet, for the new dual-blind face restoration task against other SOTA baselines.}
	\label{fig:isplunet}
\end{figure*}

\subsection{Dynamic Prior Fusion}
So far, we have established the mathematical foundation of dual-blind face restoration with ISP model, and discussed its capability to handle varying degradation levels by adjusting subspace level $k$. 
Intuitively, the best $k$ for any given $y$ would depend on its ``degradation level'', where heavier degradation calls for a smaller $k$. However, ``degradation level'' is an ambiguous concept that is difficult to define precisely, especially across different types. Therefore, finding the optimal $k$ could be technically challenging, whereas the simple workaround that enumerates every possible $k$ would incur intolerable space-time cost for practical applications.
To address this issue, we introduce a dynamic prior fusion module to
convert the discrete search of $k$ into a continuous problem, which allows flexible ensemble learning and prior utilization with a unified architecture, as shown in Fig~\ref{fig:framework}(f). Specifically, we formulate the modified ISP model with dynamic prior fusion as follows:

\begin{definition}
	A modified implicit subspace prior model with dynamic fusion module $\mathcal{F}$ is defined as follows:
	
	\begin{equation}
		\label{eqn:dynamicpriorfusion}
		\tilde{x} = G^{HQ}(E^{LQ}(y); \mathcal{F}(\mathbf{y})) = \prod_{i=0}^{n-1}G_i^{HQ}(.;\mathcal{F}_i(\mathbf{y}))\circ \prod_{j=n-1}^{0}E_j^{LQ} (y)
	\end{equation}
\end{definition}

Here $\mathbf{y} = (y_0, y_1, \dots, y_{n-1})$ denotes the collection of all subspace embedding tensors of input $y$, which are composed by the $i$-th fusion function $\mathcal{F}_i: Y_0 \times Y_1 \times \dots \times Y_{n-1} \to F_i$ to acquire the guidance map $f_i = \mathcal{F}_i(\mathbf{y})$ at level $i$. Correspondingly, the restoration mapping $G_i^{HQ}: X_i \times F_i \to X_{i+1}$ is also modified to a conditional one upon the fused guidance $f_i$ for detail replenishment over the previous layer\footnote{Although $G_i$ is a bivariate function, it can still be composed since its second parameter is fixed by the fused guidance $f_i$.}. 
Compared to the original ISP model in Eqn.~\eqref{eqn:ISP}, our modification encapsulates a wide range of powerful generative models, and further supports flexible ensemble learning of different models:

\begin{itemize}
	\item \textbf{StyleGAN} is equivalent to the case where $f_i = \phi_i(y_0)$, which corresponds to the case of a single embedding space $Y_0 = S_0$ that is shared across multiple subspaces.
	\item \textbf{U-net} introduces skip connections to convey low-level visual features into the generator, which is equivalent to $f_i = y_i$. This is the default implementation in our previous work ``HiFaceGAN''~\cite{yang2020hifacegan}.
	\item \textbf{Multi-code GAN prior~\cite{mganprior}} adopts linear fusion with weights $\bm{\alpha}$ for feature composition across channels, i.e. $\mathcal{F}_i(\mathbf{y}) =\bm{\alpha}^\top \mathbf{y}$. However, our ISPL model cascades the subspaces instead of  arranging them in parallel, which captures the inherent semantic hierarchy and enables dynamic perception-distortion tradeoff against varying degradation levels.
\end{itemize}

\subsection{Model Implementation}
\label{sec:implementation}

As shown in Fig.~\ref{fig:framework}(f), our full ISPL model involves a collaborative learning of three components: subspace embedding network $E^{LQ}$, dynamic prior fusion module $\mathcal{F}$, and restoration generator $G^{HQ}$, whose implementation details are presented below. The embedding network aims to extract informative, degradation-agnostic subspace embeddings, where the ability to discriminate between informative features and degradation artifacts is crucial. To this end, we introduce pixel-adaptive convolution\cite{PixelAdaptiveCNN} to implement our embedding function as follows (here we drop the layer index $i$ for notation simplicity):

\begin{equation}\label{eqn:embedding}
	E^{LQ}(\i) = \sum_{\j\in\Omega(\i)} \varphi(y(\j),y(\i)) w_{\Delta \j\i} y(\j)
\end{equation}

where $\i,\j$ are 2D spatial coordinates, $\Omega(\i)$ is the receptive field centering at $\i$, and $\Delta \j\i$ is the offset between $\j$ and $\i$ for indexing kernel elements. Specifically, $\varphi(\cdot,\cdot)$ measures the feature correlation between neighborhood pixels
to discriminate informative pixels from degradation-corrupted ones, and modulate the convolution kernel weights accordingly. Here we implement $\varphi(.,.)$ as a parameterized correlation metric:

\begin{equation}\label{eqn:phi}
	\varphi(y_\i,y_\j) = \mathbf{tanh}(g(y_\i)^\top 
	g(y_\j) )
\end{equation}

where $g: \mathbb{R}^C \rightarrow \mathbb{R}^D$ aims to reduce channel redundancy of the raw feature $y_\i \in \mathbb{R}^C$ via a low-dimensional projection. In practice, $g$ is implemented via a small multi-layer perceptron (MLP) to allow end-to-end training of filtering criteria, thus maximizing the selective power of the embedding network.

For the prior fusion module $\mathcal{F}$, we mainly focus on the case where $\mathcal{F}_i$ is linear to each $y_i \in \mathbf{y}$. This boils down the module architecture to a bipartite graph with $n^2$ edges, yet the total $2^{n^2}$ possibilities is still intolerable.
In this paper, we compare several representative fusion schemes in Sec.~\ref{sec:ablation} and choose the U-net variant as our benchmark implementation for its competitive performance and moderate computational cost, as shown in Fig.~\ref{fig:isplunet}.

Finally, the generative restoration mappings $G_i^{HQ}$ are implemented via spatial adaptive normalization~\cite{spade}, an effective technique for guided image synthesis. Here the initial input $x_0$ is set to $y_0$, and the final restoration result $\tilde{x}=x_n$ is defined via the following recursive formula:

\begin{equation}
	x_{i+1} = G_i^{HQ}(x_i; f_i) = \gamma_i\frac{x_i - \mu(x_i)}{\sigma_(x_i)} + \beta_i
\end{equation}

where $\mu,\sigma$ denotes mean and variance, and $\gamma_i$ and $\beta_i$ are modulation parameters inferred from fused guidance $f_i$.

\begin{figure}
	\centering
	\includegraphics[width=\linewidth]{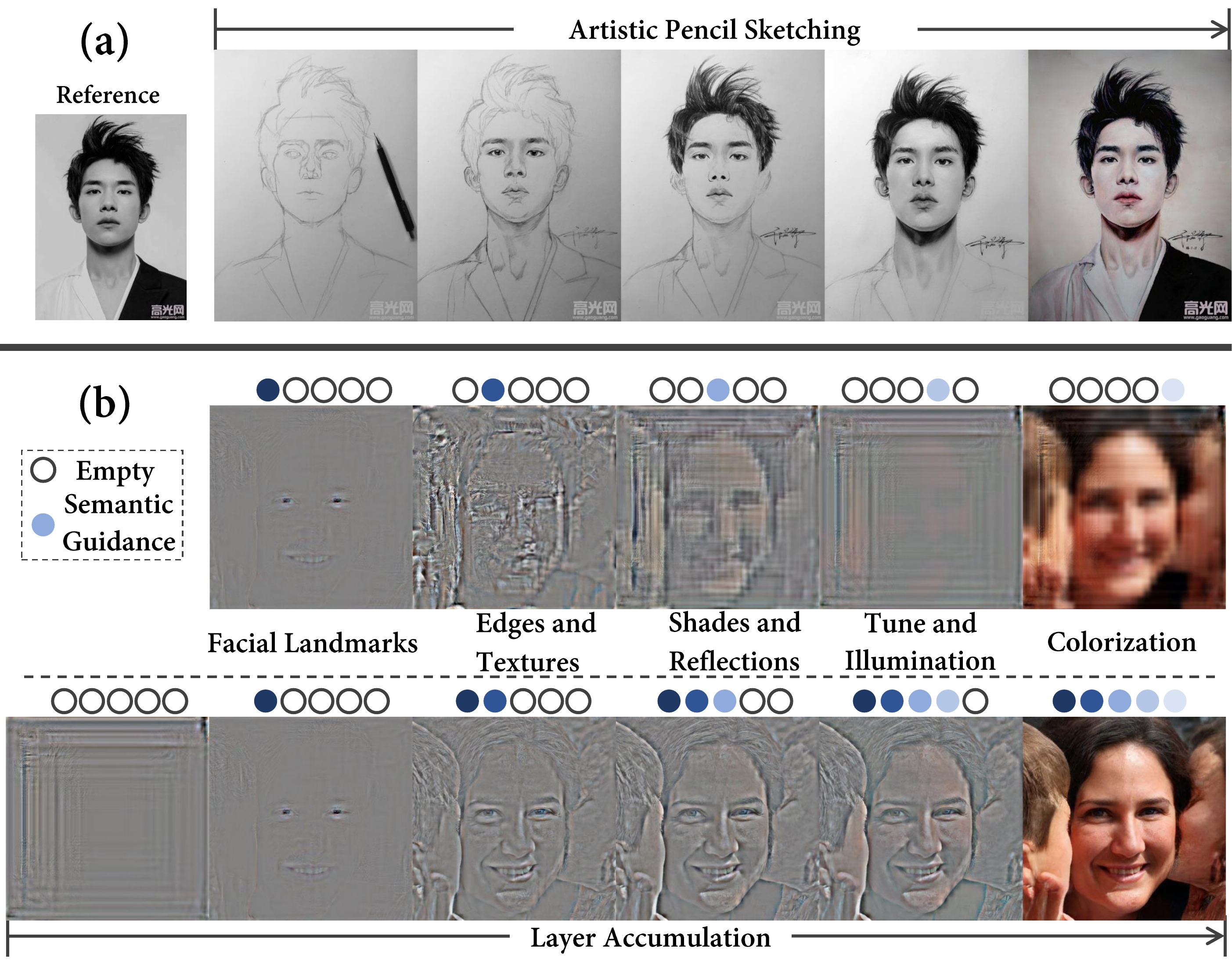}
	\caption{An illustration of the working mechanism of ISPL. (a) A step-by-step pencil sketching tutorial.
		(b) The corresponding semantic aspects learned at each subspace (top row), and the accumulated restoration procedure(bottom row).}
	\label{fig:understanding}
\end{figure}

\subsection{Learning Objective}
\label{sec:learning_obj}
Different from the objective in Eqn.~\eqref{eqn:lagrange_formulation}, ISPL does not require any hand-crafted regularization terms for training. Instead, a generic adversarial loss $\L_{GAN}$ will suffice the purpose of implicit prior learning. Here, LSGAN~\cite{lsgan} is applied for better training dynamics:
\begin{equation}\label{eqn:GAN}
	\L_{GAN} = \mathrm{E}[\|\log(D(I_{gt}) - 1\|_2^2] + \mathrm{E}[\|\log(D(I_{gen})\|_2^2]
\end{equation}

Also, multi-scale feature matching loss $\L_{FM}$~\cite{pix2pixHD} and perceptual loss $\L_{perc}$\cite{perceptual_loss} are introduced to enhance the perceptual quality and realism of localized visual details:
\begin{equation}\label{eqn:fm}
	\L(\Phi) = \sum_{i=1}^{L} \frac{1}{H_i W_i C_i} \|\Phi_i(I_{gt}) - \Phi_i(I_{gen})\|_2^2
\end{equation}

where $\Phi$ is chosen to be a multi-scale discriminator~\cite{pix2pixHD} for $\L_{FM}$, and a pretrained VGG-19 network for $\L_{perc}$.
Finally, combining Eqn.~\eqref{eqn:GAN} and~\eqref{eqn:fm} leads to the full learning objective:
\begin{equation}\label{eqn:recon}
\L_{full} = \L_{GAN} + \lambda_{FM}\L_{FM} + \lambda_{perc}\L_{perc}
\end{equation}

\textbf{Remarks} Existing image restoration works~\cite{srcnn}\cite{vdsr}\cite{edsr} typically adopt a reconstruction loss $\L_{recon} = \|\tilde{x} - x\|$ to minimize distortion, which often leads to severe drop in perceptual quality and insufficient amount of details~\cite{esrgan}. Instead, we lift the constraint on signal distortion and adopt a purely perceptual-oriented learning objective, which usually implies a PSNR drop of up to 3 db~\cite{Tradeoff}. Yet our empirical results (Table~\ref{tab:ffhq_sota}) over multiple restoration subtasks show significant PSNR increase over SOTA baselines, which leads to questions of P-D tradeoff for category-specific restoration tasks. The corresponding analysis is detailed in Sec.~\ref{sec:pd_tradeoff}

\subsection{Subspace Visualization}
\label{sec:understanding}

To better illustrate the mechanism of ISPL and justify our algorithmic design, we visualize the subspace embeddings of ISPL over a 16x downsampled face. Here we implement the fusion mappings $\left\{\mathcal{F}_i\right\}$ with skip-connections to preserve the original semantic prior of learned subspaces, and isolate each individual subspace by replacing the fusion guidance in other layers (blue dots) with constant tensors $C=0.5$ (hollow circles), leading to a plain gray background to better contrast the embedding contents (Fig.~\ref{fig:understanding}(b), top row).
Moreover, to justify our heuristics on hierarchical subspace decomposition, we compare the accumulative restoration results of ISPL (Fig.~\ref{fig:understanding}(b), bottom row) with a step-by-step portrait sketching tutorial (Fig.~\ref{fig:understanding}(a)) from professional artists. Our restoration algorithm shares an astonishing resemblance with the actual art creation process, where the initial common subspace is captured through high-level structural components, such as facial contour and landmarks. Thereafter, fine-grained textures are depicted upon the base structure, following with contrast enhancement, global illumination and colorization. This indicates that our ISPL algorithm can successfully model the content prior of real-world images via interpretable visual semantic decomposition onto separate subspaces, allowing progressive enhancement of perceptual quality and realism in a semantic-guided fashion, all accomplished without explicit prior guidance.


\section{Experiments}
In this section, we evaluate the proposed ISPL on a wide range of restoration tasks. First, we perform an ablation study in Sec~\ref{sec:ablation} to justify the implicit formulation and determine the optimal implementation. Afterwards, Sec.~\ref{sec:against_SOTA} reports the benchmark performance of ISPL over individual degradation types, and Sec.~\ref{sec:FaceRenov} studies the generalization and adaptability across synthetic and real-world degradation domain. Afterwards, we revisit the theory of perception-distortion tradeoff in Sec.~\ref{sec:pd_tradeoff} to explain ISPL's superior performance, and demonstrates possible extensions of ISPL towards natural images and videos in \ref{sec:extension}.

\begin{figure}
	\centering
	\includegraphics[width=\linewidth]{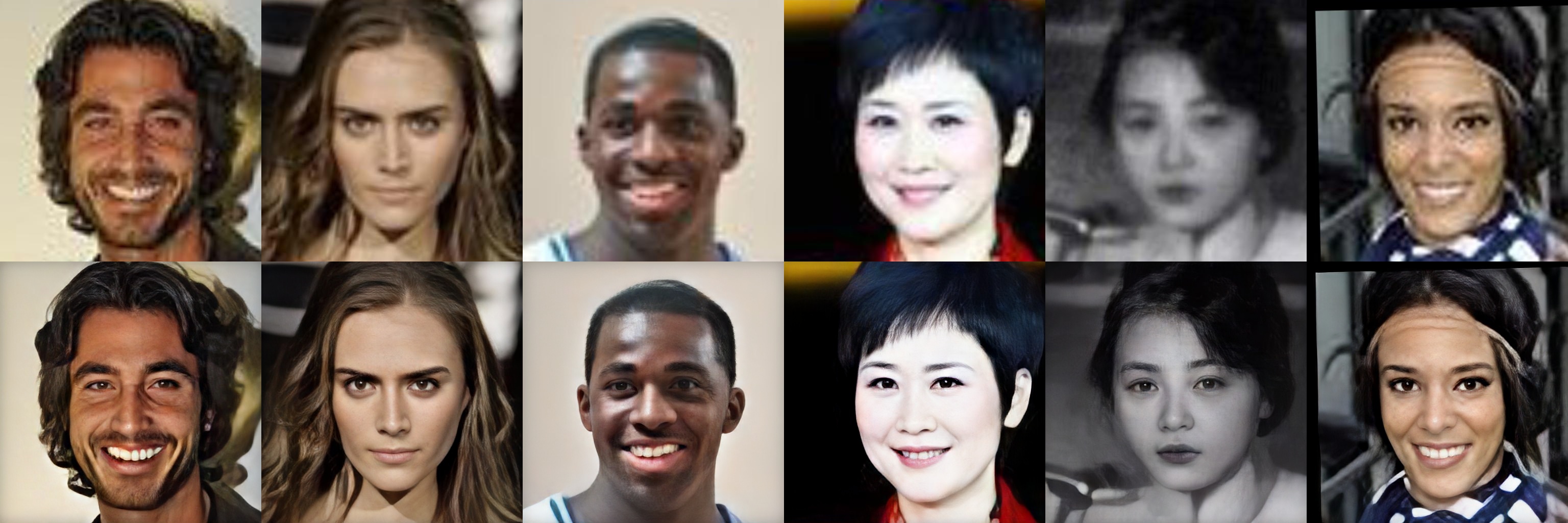}
	\caption{Examples of the FaceRenov dataset. The bottom row displays pseudo HQ labels generated from top row LQ inputs with commercial softwares.}
	\label{fig:facerenov}
\end{figure}

\subsection{Experimental Settings}
\label{sec:exp_setting}

\textbf{Data Collection.}
\label{sec:facerenov} 
We adopt the challenging FFHQ~\cite{stylegan_ffhq} dataset to fully reflect the challenge in real-world face restoration, which are resized to $512\times 512$ and perturbed under various degradation settings for different tasks. Besides synthetic degradation tests in~\cite{yang2020hifacegan}, we further develop a real-world degradation benchmark dataset ``FaceRenov'' to help quantify the degradation domain gap and generalization ability of different restoration methods. As it is often difficult to acquire the underlying HQ image for real-world LQ faces, we ask 5 professional photographers to compare the results of several AI-powered face repairing/retouching softwares (see Appendix~\ref{sec:appendix-apps} for the list), and select the best one as pseudo HQ label. Fig.~\ref{fig:facerenov} showcases several sample pairs from the dataset. Overall, the software-processed HQ labels is visually plausible with relatively decent amount of details.

\textbf{Evaluation Protocols.}
\label{sec:quantification}
The objective for real-world restoration task typically depends on the particular scenario: For video analysis and surveillance applications, the semantic fidelity, especially the identity of restored faces, is the top priority; For photo retouching and face beautification applications, the perceptual and aesthetic quality should be more emphasized, while a slight content discrepancy is often allowed. In this paper, we perform a comprehensive evalution in three aspects: statistical accuracy as measured by PSNR and SSIM\cite{MultiscaleSSIM}, semantic fidelity that includes the feature embedding distance (FED) and landmark localization error (LLE) with a pretrained face recognition model~\cite{dlib_paper}, and perceptual realism, including FID~\cite{FID} and LPIPS~\cite{LPIPS} for distributional and pairwise distance between original and generated samples. Also, the non-reference NIQE~\cite{niqe} metric is introduced to evaluate in-the-wild face images. 

For the new FaceRenov dataset, the domain gap between real-world and synthetic degradation is quantified by comparing the training/testing performance across datasets. Specifically, each baseline is evaluated under three different protocols: S2S (synthetic-to-synthetic), S2R(synthetic-to-real) and R2R (real-to-real). The degradation gap is defined via the relative performance drop from S2S to S2R mode, and the generalization ability is characterized by the relative gain from S2R to R2R. The quantification results are analyzed in Sec.~\ref{sec:FaceRenov}.

\textbf{Training and Testing Details.} We adopt a 5-layer ISP model with a 2x bilinear sampling across adjacent layers, so the innermost subspace embedding $y_0$ is of size $16\times16$. Both the encoder and generator have a base filter number of $c=64$, and is doubled after each layer (up to $1024$). The loss weights $\lambda_{FM}$ and $\lambda_{perc}$ are both set to 10.0.
All models are trained with Adam optimizer with $lr=0.0002$ and $\beta = (0.9, 0.999)$, with constant rate for the first 30 epochs, followed by linear decay of 20 epochs. This takes roughly 2~3 days for training on a single Nvidia V100 card.
For benchmark comparison, all baseline methods are retrained over synthetic FFHQ and FaceRenov following the official implementation unless training code is not available~\footnote{Three exceptions are: GFRNet~\cite{BlindFR-ECCV2018}, RIDNet~\cite{RIDNet} and EPGAN~\cite{EPGAN}, which is then evaluated with author-released models.} For super resolution methods, the input are downscaled accordingly via bicubic interpolation.

\begin{table}[!t]
	\renewcommand{\arraystretch}{1.2}
	\centering
	\caption{Ablation study results on 16x face hallucination.}
	\label{tab:ablation}
	\resizebox{0.45\textwidth}{!}{
		\begin{tabular}{|c|cc|ccc|cc|}
			\hline
			Prior Type & \multicolumn{2}{c|}{\textbf{Explicit}} & \multicolumn{3}{c|}{\textbf{Implicit, Sparse}} & \multicolumn{2}{c|}{\textbf{Implicit, Dense}}\\
			Metrics & EP-map & EP-16x & ISP-Y0 & ISP-unet & ISP-unet & ISP-concat & ISP-matrix \\
			has PAC & - & - & \cross & \cross & \check & \check & \check \\
			\hline
			PSNR $ \uparrow $ & 20.968 & 23.541 & 23.651 & 23.689 &  23.705 & 23.852 & \textbf{23.934}\\
			SSIM $ \uparrow $ & 0.596 & 0.610 & 0.615 & 0.616 & 0.619 & 0.629 & \textbf{0.631}\\
			MS-SSIM $ \uparrow $ & 0.718 & 0.811 & 0.818 & 0.817 & 0.819 & 0.820 & \textbf{0.821}\\
			\hline
			FED $ \downarrow $ & 0.4595 & 0.3296 & 0.3236 & 0.3212 & \textbf{0.3182} & 0.3409 & 0.3387\\
			LLE $ \downarrow $ & 4.143 & 3.227 & 3.157 & 3.152 & 3.137 & 2.994 & \textbf{2.952}\\
			\hline
			FID $ \downarrow $ & 52.701 & 14.365 & 13.154 & 11.403 & \textbf{11.389} & 17.485 & 17.206\\
			LPIPS $ \downarrow $ & 0.3967 & 0.2609 & 0.2467 & 0.2462 & \textbf{0.2449} & 0.2808 & 0.2793\\
			NIQE $ \downarrow $ & 10.367 & 7.446  & 7.011 & 7.000 & \textbf{6.767} & 8.389 & 7.545 \\
			\hline
		\end{tabular}
	}
\end{table}

\subsection{Ablation Study}
\label{sec:ablation}
In this section, we perform an ablation study over the most challenging 16x face hallucination task to validate the necessity of implicit subspace prior formulation, and explore the possible architectural designs for ISPL.
Note that our ablation design is by no means exhaustive, as the total number of possible configurations ($O(2^{n^2})$) is too overwhelming. Instead, we aim to reveal some general rules by comparing the following three cases:

\begin{itemize}
	\item \textbf{Explicit Prior} utilizes the original SPADE~\cite{spade} architecture and fix subspace embeddings to be either face parsing maps extracted from~\cite{MaskGAN} or the raw input with 16-pixel mosaics, denoted as \emph{EP-map} and \emph{EP-16x}, respectively.
	\item \textbf{Implicit Sparse Prior} only activates one pathway for each fusion function $\mathcal{F}_i$, where \emph{ISP-Y0} assumes $\mathcal{F}_i = y_0$ and \emph{ISP-unet} assumes $\mathcal{F}_i = y_i$. Feature maps are spatially aligned with bilinear interpolation. For u-net variant, we further compare the efficacy of pixel adaptive convolution (PAC) over traditional convolution operator.
	\item \textbf{Implicit Dense Prior} activates all $n$ pathways for each fusion function, where \emph{ISP-concat} directly concatenates spatially-aligned embeddings via $\mathcal{F}_i = [y_0;y_1;\dots;y_{n-1}]$, and \emph{ISP-matrix} performs linear fusion via $\mathcal{F}(\mathbf{y}) = \mathbf{Wy}$, where $\mathbf{W} \in \mathbb{R}^{n\times n}$ is a learnable fusion matrix. For matrix fusion, the channel numbers of each embedding is aligned to $128$ with $1\times1$ convolution.
\end{itemize}

As shown in Table~\ref{tab:ablation}, implicit prior models completely dominate explicit ones, especially over the \emph{EP-map} case with semantic parsing maps as guidance. This verifies the superiority of our implicit formulation in restoration tasks.
For sparse prior, the U-net variant with multi-stage prior fusion works better than its single-stage counterpart ISP-Y0, and the pixel-adaptive embedding further improves the feature discrimination ability. For dense prior based methods, a higher statistical score is achieved at the cost of perceptual quality, where the matrix fusion scheme outperforms direct concatenation due to the end-to-end fusion weight learning. We note that a better fusion scheme could be achieved by further exploring the combination of sparse and dense options with neural architecture search, but that is quite beyond the scope of this paper. 
In the following sections, we shall settle for the pixel-adaptive ISP-unet variant (as in Fig.~\ref{fig:isplunet}) as our benchmark solution due to its relatively high performance and moderate memory cost, which is more than adequate for our task at hand.

\begin{table*}
	\renewcommand{\arraystretch}{1.1}
	\centering
	\caption{Quantitative comparison on synthetic FFHQ dataset. The best result for each metric is shown in bold format. (Up arrow means the higher score is preferred, and vice versa.)}
	\label{tab:ffhq_sota}
	\begin{tabular}{|c|c|ccc|cc|ccc|}
		\hline
		& & & Statistical & &\multicolumn{2}{c|}{Semantic} & \multicolumn{3}{c|}{Perceptual}\\
		Task & Methods & PSNR $ \uparrow $ &  SSIM $ \uparrow $ & MS-SSIM $ \uparrow $ & FED $ \downarrow $ & LLE $ \downarrow $ & FID $ \downarrow $ & LPIPS $ \downarrow $ & NIQE $ \downarrow $ \\
		\hline
		& EDSR~\cite{edsr} & 30.188 & 0.824 & 0.961 & 0.0843 & \textbf{2.003} & 20.605 & 0.2475 & 13.636 \\
		& SRGAN~\cite{srgan} & 27.494 & 0.735 & 0.935 & 0.1097 & 2.269 & 4.396 & 0.1313 & 7.378 \\
		\textbf{Face Super} & ESRGAN~\cite{esrgan} & 27.134 & 0.741 & 0.935 & 0.1107 & 2.261 & 3.503 & 0.1221 & 6.984 \\
		\textbf{Resolution}  & SRFBN~\cite{feedback_sr} & 29.577 & 0.827 & 0.953 & 0.0984 & 2.066& 20.032 & 0.2406 & 13.901 \\
		(4x, Bicubic)& Super-FAN~\cite{SuperFAN} & 25.463 & 0.729 & 0.913 & 0.1416 & 2.333 & 14.811 & 0.2357 & 8.719 \\
		& WaveletCNN~\cite{waveletcnn_conf} & 28.750 & 0.806 & 0.952 & 0.0964 & 2.072 & 16.472 & 0.2443 & 12.217 \\
		& \textbf{ISPL(ours)} & \textbf{30.824} & \textbf{0.838} & \textbf{0.971} & \textbf{0.0716} & 2.071& \textbf{1.898} & \textbf{0.0723} & \textbf{6.961} \\
		\hline\hline
		
		& Super-FAN & 20.536 & 0.540 & 0.744 & 0.4297 & 4.834& 63.693 & 0.4411 & 7.444 \\
		\textbf{Hallucination} & ESRGAN & 21.001 & 0.576 & 0.697 & 0.5138 & 5.902 & 50.901 & 0.3928 & 15.383 \\
		(16x, Mosaic) &WaveletCNN & \textbf{23.810} & \textbf{0.675} & \textbf{0.837} & 0.3713 & 3.729 & 60.916 & 0.4909 & 11.450 \\
		& \textbf{ISPL(ours)} & 23.705 & 0.619 & 0.819 & \textbf{0.3182} & \textbf{3.137} & \textbf{11.389} & \textbf{0.2449} & \textbf{6.767} \\
		\hline\hline
		
		\textbf{Denoising} & RIDNet~\cite{RIDNet} & 25.432 & 0.731 & 0.891 & 0.2128 & 2.465& 36.515 & 0.3864 & 13.002 \\
		(1/3 Gaussian, & WaveletCNN & 26.530 & 0.754 & 0.895 & 0.2441 & 2.728 & 26.731 & 0.3119 & 11.395 \\
		1/3 Poisson, & VDNet~\cite{VDNet} & 27.718 & 0.797 & 0.928 & 0.1551 & 2.297 & 15.826 & 0.2458 & 14.262 \\
		1/3 Laplacian) & \textbf{ISPL(ours)} & \textbf{31.828} & \textbf{0.845} & \textbf{0.957} & \textbf{0.1109} & \textbf{2.090} & \textbf{3.926} & \textbf{0.0868} & \textbf{7.341} \\
		\hline\hline
		
		\textbf{Deblurring} & DeblurGAN~\cite{DeblurGAN} & 25.304 & 0.718 & 0.894 & 0.1786 & 3.219 & 14.331 & 0.2574 & 12.697 \\
		(1/2 Motion blur& DeblurGANv2~\cite{DeblurGANv2} & 26.908 & 0.773 & 0.913 & 0.1043 & 3.036 & 10.285 & 0.2178 & 13.729 \\
		1/2 Gaussian blur)& \textbf{ISPL(ours)} & \textbf{28.928} & \textbf{0.793} & \textbf{0.954} & \textbf{0.0913} & \textbf{2.156} & \textbf{2.580} & \textbf{0.0874} & \textbf{7.426} \\
		\hline\hline
		
		& ARCNN~\cite{ARCNN} & \textbf{33.021} & 0.879 & 0.972 & 0.0845 & \textbf{1.959} & 9.761 & 0.1551 & 14.827 \\
		\textbf{JPEG artifact} & EPGAN~\cite{EPGAN} & 32.780 & \textbf{0.882} & \textbf{0.976} & \textbf{0.0814} & 1.979 & 10.250 & 0.1638 & 13.729 \\
		\textbf{removal} & \textbf{ISPL(ours)} & 31.611 & 0.850 & 0.970 & 0.0842 & 2.057& \textbf{1.880} & \textbf{0.0541} & \textbf{6.911} \\
		\hline\hline
		
		& Degraded Input & 22.905 & 0.465 & 0.756 & 0.2875 & 3.936 & 63.670 & 0.6828 & 21.955 \\
		& Super-FAN & 24.818 & 0.549 & 0.818 & 0.2495 & 3.705 & 32.800 & 0.4283 & 12.154 \\
		& ESRGAN & 24.197 & 0.564 & 0.816 & 0.2761 & 3.771 & 28.053 & 0.4141 & 11.382 \\
		\textbf{Dual-blind Restoration} & WaveletCNN & 24.404 & 0.648 & 0.817 & 0.2821 & 3.690 & 58.901 & 0.3102 & 15.530 \\
		(Full Degradation) & DeblurGANv2 & 23.704 & 0.494 & 0.776 & 0.2403 & 4.412 & 49.329 & 0.6496 & 21.983 \\
		& ARCNN & 24.187 & 0.539 & 0.787 & 0.2580 & 3.833 & 60.864 & 0.6424 & 18.880 \\
		& GFRNet~\cite{BlindFR-ECCV2018} & 25.227 & \textbf{0.686} & 0.854 & 0.2524 & 3.371 & 48.229 & 0.4591 & 20.777 \\
		& \textbf{ISPL(ours)} & \textbf{25.837} & 0.674 & \textbf{0.881} & \textbf{0.2055} & \textbf{2.701} & \textbf{8.013} & \textbf{0.2093} &\textbf{7.272} \\
		\hline\hline
		
		\textbf{FFHQ GT} & --- & $+\infty$ & 1 & 1 & 0 & 0& 0 & 0 & 7.796 \\
		\hline
		
	\end{tabular}
\end{table*}

\subsection{Comparison with Existing Restoration Methods}
\label{sec:against_SOTA}
We benchmark the performance of ISPL over six restoration subtasks against corresponding baselines, with the following synthetic degradation prior setting (the degradation script is provided along with our code):
\begin{itemize}
	\item \textbf{Face Super Resolution} with 4x bicubic downsampling.
	\item \textbf{Hallucination} with 16 pixel mosaics.
	\item \textbf{Denoising} for Gaussian, Poisson and Laplacian noises
	\item \textbf{Deblurring} for both Gaussian and motion blur kernels
	\item \textbf{JPEG artifact removal} with quality $Q\in[50,85]$.
	\item \textbf{Dual-blind Restoration} with a hybrid degradation of all previous subtypes (except 16x mosaic).
\end{itemize}

Quantitative results are shown in Table~\ref{tab:ffhq_sota}. ISPL achieves dominant perceptual performance over all baselines, with 3-10 times gain on FID and 50\%-200\% gain on LPIPS over all baselines, indicating an  extreme efficacy in capturing complex content prior. Furthermore, ISPL is capable of capturing the holistic prior of natural images instead of just focusing on facial details, leading to even better natural image statistics (NIQE) than the original FFHQ dataset. For statistical performance, ISPL is also superior in learning pixel-adaptive noise filters and replenishing fine-grained image contents, leading to 4.11 db gain on PSNR over state-of-the-art denoising methods. Overall, the versatility of proposed implicit subspace prior model for face restoration is well justified with superior performances on a broad range of restoration subtasks.

\subsection{Quantifying the Domain Gap and Generalizability}
\label{sec:FaceRenov}

\begin{figure}
	\centering
	\includegraphics[width=\linewidth]{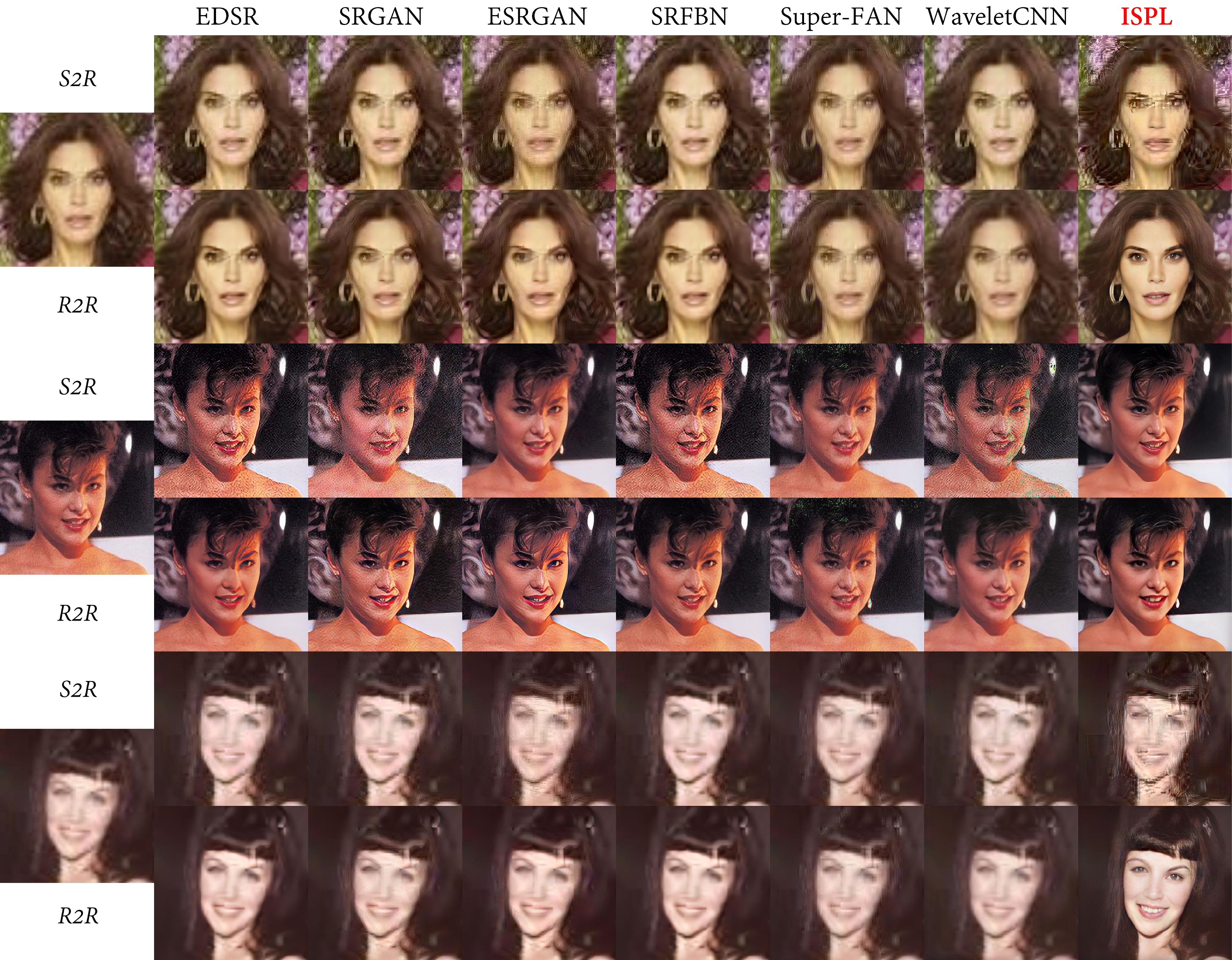}
	\caption{Restoration performance on real-world face images. As the degradation becomes heavier, models trained on synthetic data (including ours) deteriorate rapidly, while our model trained on FaceRenov consistently deliver high quality restoration results.}
	\label{fig:casia}
\end{figure}

We analyze the degradation domain gap and generalization capability of existing super resolution models in Table.~\ref{tab:face_renov}. Here we use the average changing rate on perceptual metrics (FID and LPIPS) to monitor the prior divergence between testing data and learned models. Overall, all methods suffer from intensive performance drop when transferring to the FaceRenov dataset (S2R), which is relieved after training with the same dataset (R2R). In particular, we note the adversarial effect of domain shift is closely correlated to generative capacity, where GAN-based methods exhibit 10-13 times increase on perceptual divergence when shifting from S2S to S2R case, as opposed to 2-3 times for non-GAN methods. Similarly, GAN-based methods also better adapt to realistic degradation in FaceRenov dataset, with an average of 43.6\% performance gain after retraining on FaceRenov. To further illustrate the distinction between our ISPL and different methods, we compare the restoration results over real-world LQ images in Fig.~\ref{fig:casia}, where the synthetic degradation is assumed to be 4x bicubic. Apparently, models trained under synthetic prior assumption fail to cope with additive noise, leading to amplified noise artifacts in S2R test. After aligning the degradation prior across training and testing, all baselines can successfully identify and reduce noise artifacts, but still lacks the ability to replenish fine-grained facial details. In contrast, our ISPL algorithm is much more active in detail replenishment, as evidenced by more intense artifacts with inaccurate semantic embeddings (S2R case) and stunning visual quality and detail realism against severely corrupted images (R2R case). In conclusion, our implicit prior learning scheme can better adapt to complex degradation and content prior by dynamically replenishing the missing details in the degraded input, leading to consistently high-quality restoration performance over real-world face images.

\begin{table}[!pt]
	\renewcommand{\arraystretch}{1.2}
	\centering
	\caption{Quantified domain gap and generalization of different super resolution methods. Each measure is evaluated twice under S2R and R2R protocols.}
	\label{tab:face_renov}
	\begin{tabular}{|m{0.09\textwidth}|m{0.08\textwidth}m{0.08\textwidth}|m{0.06\textwidth}m{0.06\textwidth}|}
		\hline
		Methods &  FID $ \downarrow $ & LPIPS $ \downarrow $ & S2R gap($\times$) & R2R Gain(\%) \\
		\hline
		EDSR &  82.172/69.717 & 0.408/0.349 & 2.8 & 14.8 \\
		SRGAN & 73.960/36.415 & 0.362/0.240 & 9.8 & 42.2 \\
		ESRGAN &  79.833/37.205 & 0.376/0.241 & 12.9 & 44.1 \\
		SRFBN &   75.095/66.950 & 0.347/0.331 & 2.5 & 7.7 \\
		Super-FAN &  66.296/51.034 & 0.381/0.272 & 3.0 & 12.9 \\
		WaveletCNN & 83.739/54.595 & 0.402/0.321 & 3.3 & 27.5\\
		\textbf{ISPL(ours)} & \textbf{44.352/22.467} & \textbf{0.251/0.152} & \textbf{13.4} & \textbf{44.4}\\
		\hline
	\end{tabular}
\end{table}


\subsection{Improving the Perception-Distortion Tradeoff}
\label{sec:pd_tradeoff}

The classic P-D tradeoff theory~\cite{Tradeoff} suggests the existence of a performance boundary over the P-D plane that is monotone and convex regardless of the restoration algorithm. This implies that perception and distortion are inherently at odds with each other, and there is no way to optimize one criteria indefinitely without compromising the other. Naturally, the following question arises:
How far away are we still from reaching the theoretical lower bound, and what would be the best approach to achieve that?

\begin{figure*}
	\centering
	\begin{minipage}{\linewidth}
		\includegraphics[width=0.48\linewidth]{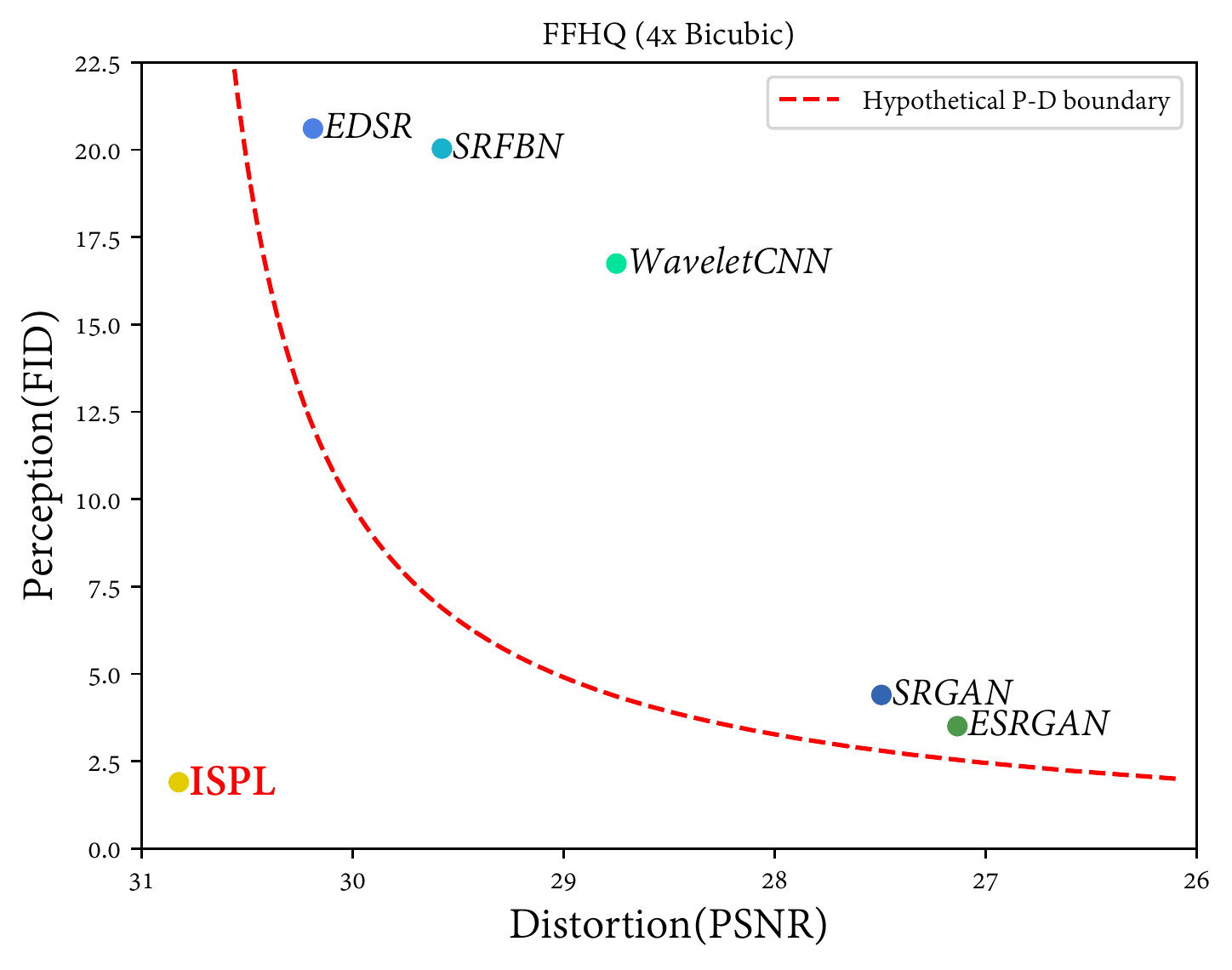}
		\includegraphics[width=0.48\linewidth]{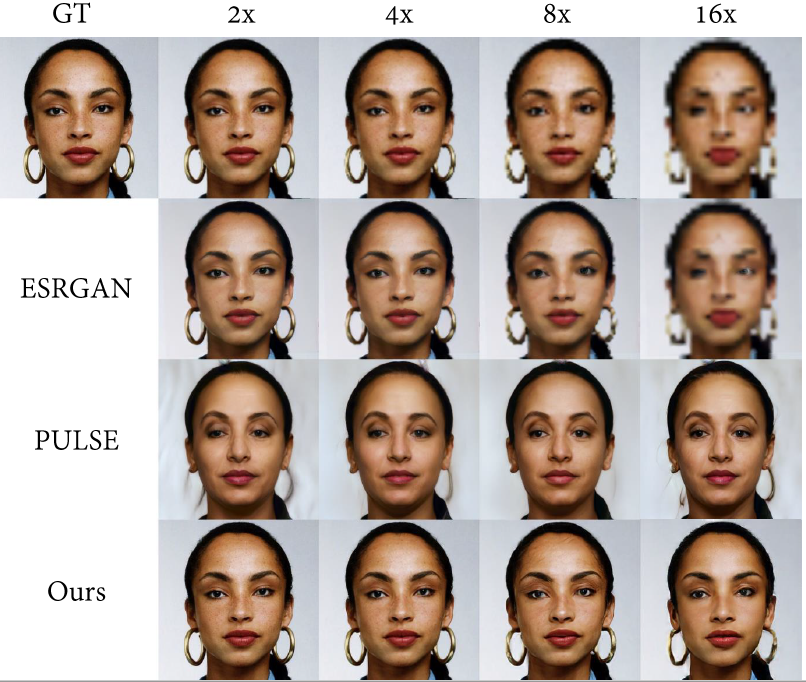}	
	\end{minipage}
	\caption{\textbf{Dynamic Perception-Distortion Tradeoff.} \emph{Left}: The performances at 4x bicubic super resolution and blind restoration task on FFHQ dataset. ISPL dominates existing methods simultaneously on both aspects that it even reach beyond our initial hypothesis on the P-D boundary. \emph{Note the PSNR axis is inverted to better illustrate the tradeoff relationship.} \emph{Right}: Super resolution results under varying scaling factors. Compared to explicit prior models, ISPL consistently delivers high-quality restoration results with rich facial details.}
	\label{fig:pd_tradeoff}
\end{figure*}

Let us first visualize the P-D performance on 4x face super resolution (SR) in Fig.\ref{fig:pd_tradeoff} (left column). On the first glance, it seems that existing SR researches are on the verge of an inevitable tradeoff, as each method can only surpass other competitors on one criteria, while being outperformed on the other. As such, one might be tempted to extrapolate a hypothetical boundary (as shown in red dashed curve) below the observed data points. We shall notice, however, that the observed tradeoff does not necessarily imply proximity of existing methods to the theoretical limit --- our ISPL can surpass existing SR methods simultaneously on both metrics and push forward the P-D boundary far beyond our initial hypothesis. This poses a rather puzzling question towards the understanding of P-D tradeoff, as well as the efficacy of existing SR techniques for restoration tasks.

A key to understand this puzzle is to realize that the P-D boundary is not only subject to the degradation, but also the ``intrinsic complexity'' of the underlying image contents. In this regard, face images, being a subset of all natural images, would have a much lower boundary than a generic estimation based on natural image statistics. Yet to reach this bound one must learn how to capture and utilize the corresponding prior from observed data.
(This is quite similar to data compression, although the generic bitrate limit is given by Shannon entropy, one can easily go beyond this limit for a specific data source by exploiting its unique patterns.)
Empirically, existing research efforts on SR are mostly targeted at a single criteria, where the progress in overall P-D performance is more or less stagnant. On the other hand, our ISPL provides a principled approach to exploit complex prior and improve the P-D performance simultaneously on both metrics. 

\textbf{Exploring Dynamic P-D Tradeoff}
Besides performance improvement, ISPL also provides an enhanced P-D tradeoff exploration scheme for varying testing conditions that requires no fine-tuning of model parameters.
Fig.\ref{fig:pd_tradeoff} (right column) shows the restoration results of different SR methods over an image downscaled $2,4,8,16$ times, respectively. Two baselines assuming either fixed degradation (ESRGAN) or content prior (PULSE) fail to handle varying testing cases, while our ISPL can automatically adjust the restoration strategy against different inputs:
For mildly degraded images, ISPL can well preserve fine-grained visual details (such as freckles) with minimal signal distortion, and as the degradation intensifies, the restoration strategy seamlessly shifts to the perceptual side by hallucinating plausible facial contents based on partial semantics in the degraded input. Consequently, ISPL delivers consistent high-quality restoration results regardless of the input quality.

\subsection{Extension to General Cases}
\label{sec:extension}
Although mainly focused on face restoration, ISPL does not rely on any facial prior assumption, thus is equally effective for more generalized problem setting. Here we demonstrate ISPL's versatility over two related low-level vision tasks.

\begin{figure}
	\centering
	\includegraphics[width=\linewidth]{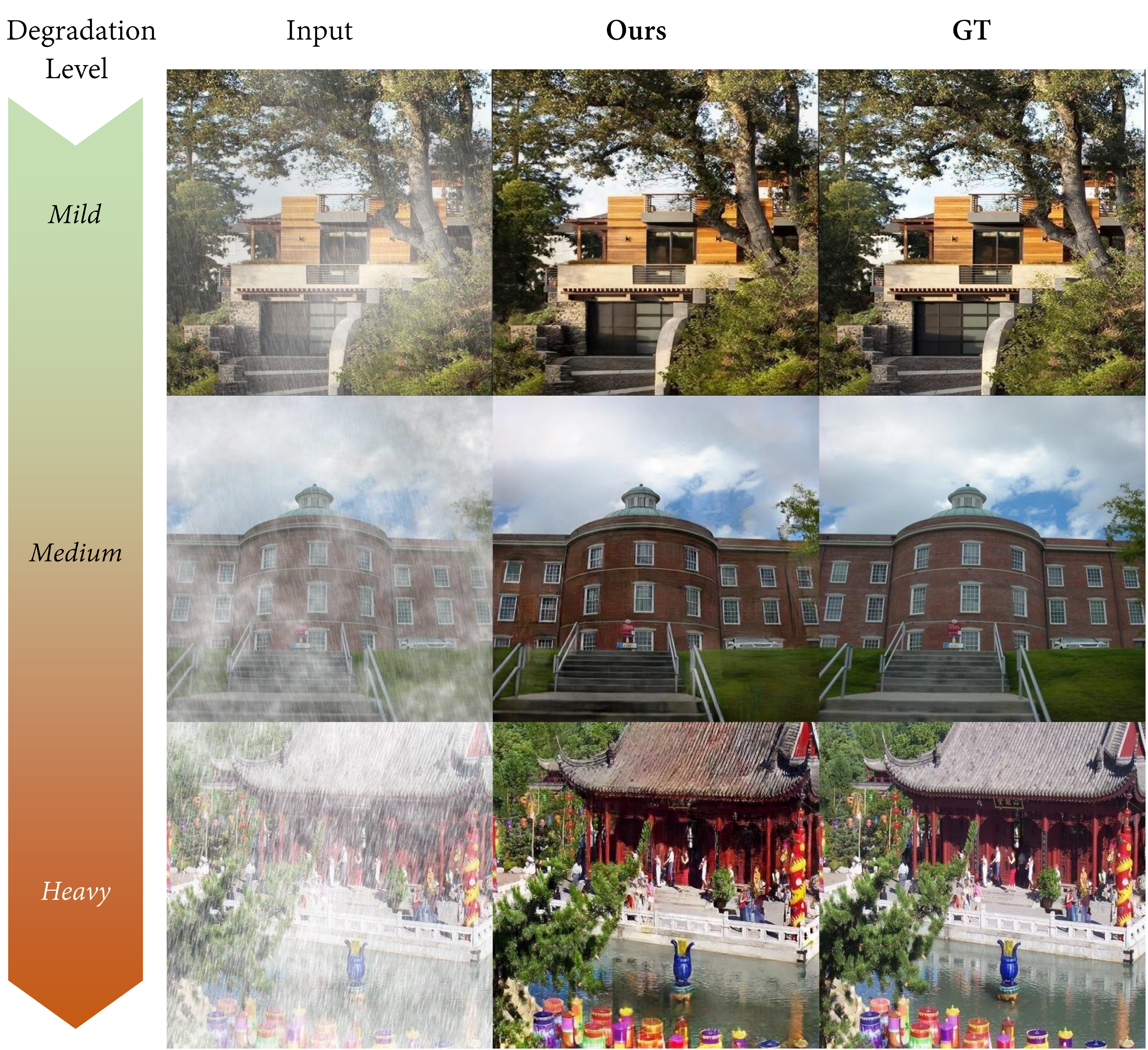}
	\caption{Extension of ISPL to joint rain streak and haze removal of complex natural scene images.}
	\label{fig:real-world}
\end{figure}

\textbf{Natural Images with Inhomogeneous Degradation.}  Fig.~\ref{fig:real-world} shows the restoration results of three natural scene images with rain streak and haze of varying intensity, where ISPL consistently produces high quality restoration results against all degradation levels. Note in the last case the rooftop is recolored to brown, which is probably due to the heavy overcast of rain streak and haze over the original content. Nevertheless, the tile textures are still well preserved, indicating that ISPL still manages to capture correct semantic and structure details. This again demonstrates the versatility of the proposed implicit formulation.

\begin{figure}
	\centering
	\includegraphics[width=\linewidth]{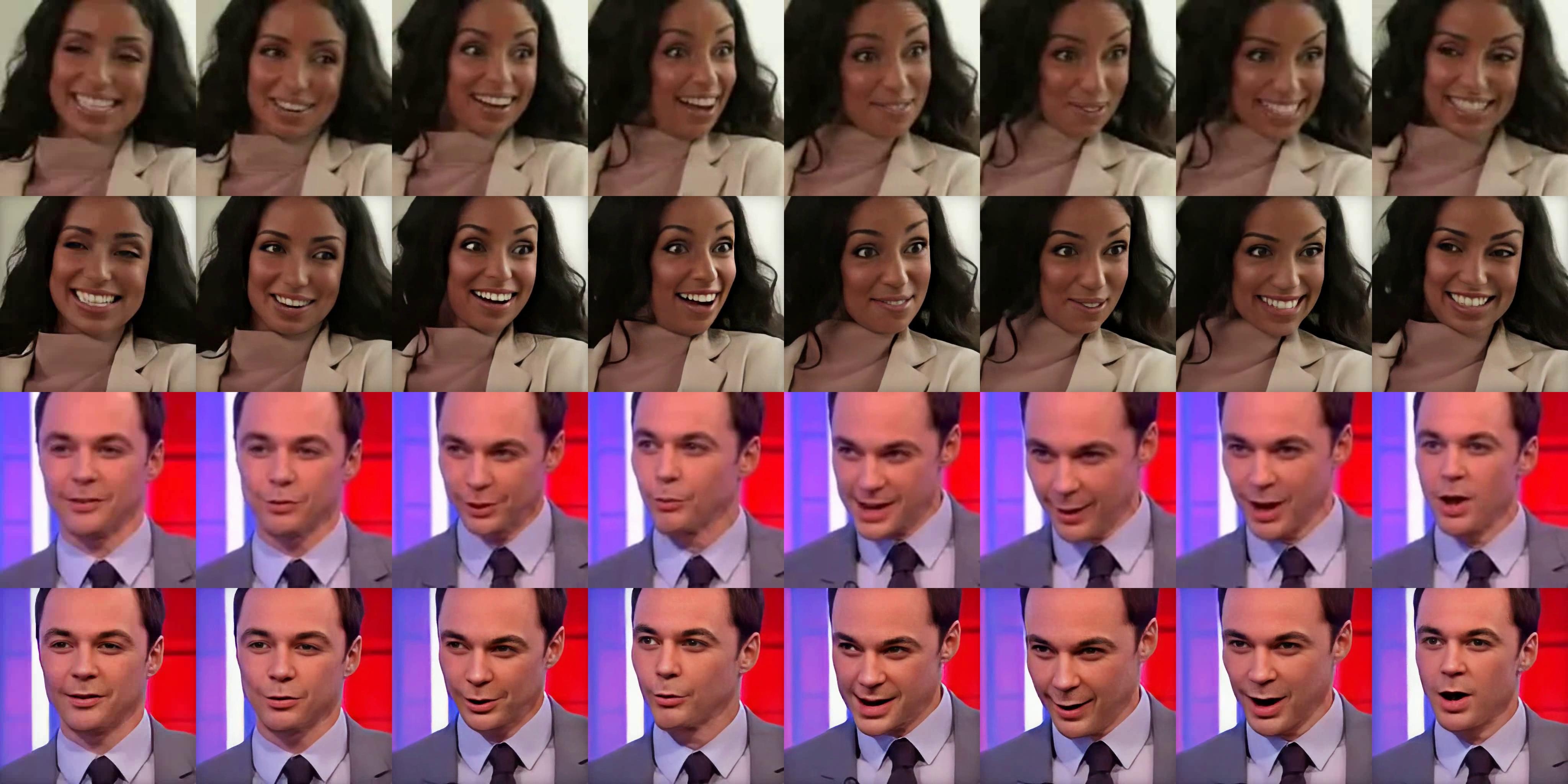}
	\caption{Video restoration results via ISPL (trained with FaceRenov). Row 1/3: input; Row 2/4: restored results.}
	\label{fig:video}
\end{figure}

\textbf{Video Restoration.} Apart from the superiority in image restoration, ISPL is also temporally stable when applied to consecutive frames of talking head videos, as shown in Fig.~\ref{fig:video}\footnote{More video results are hosted at \url{https://github.com/Lotayou/Face-Renovation-teaser-gifs}.}. This is mainly due to the pixel-adaptive feature filtering and normalization schemes, which effectively removes flickering noises in the source video to ensure spatial-temporal consistency across restored frames.


\section{Conclusion}
Face restoration is an inherently ill-posed problem, where existing methods typically enforce explicit prior regularization to relieve such a pathology, but often at the expense of limited generation ability towards real-world applications. In this paper, we investigate the more practical ``dual-blind'' face restoration problem, which aims to capture and utilize prior knowledge without explicitly modeling it, thus leading to a generic-purpose solution to face restoration under arbitrary degradation. The resulted solution, termed as ``implicit subspace prior learning'' (ISPL), adopts an implicit formulation to connect LQ and HQ domain via a common embedding space, and allows effective joint degradation and content prior learning in an end-to-end fashion.
To better cope with complex testing conditions, a subspace decomposition module is proposed to capture the structural variation of the common embedding space by adjusting the level of shared subspaces. Moreover, a dynamic prior fusion module is designed to convert the discrete subspace architecture search problem into a continuous learning one,
leading to a more advanced dynamic P-D tradeoff scheme that requires no fine-tuning at testing stage. Extensive experiments on both synthetic and real-world degraded dataset have proved the significant performance gain of our method, and justified the superiority of our ISPL formulation over explicit ones. 
With the emergence of deep neural networks, we no longer hand-craft our restoration models, and we hope this work can inspire a new research paradigm for low-level vision tasks that doesn't require hand-crafted prior models either.

\ifCLASSOPTIONcompsoc
\section*{Acknowledgments}

This work is done when Lingbo Yang is interning at DAMO Academy, Alibaba Group, and supported by the National Natural Science Foundation of China (61931014, 61632001) and High-performance Computing Platform of Peking University, which are gratefully acknowledged. We would like to thank Chang Liu for his constructive comments.

\else
\section*{Acknowledgment}
\fi

\appendices
\section{List of Face Restoration Applications}
\label{sec:appendix-apps}
For FaceRenov, three different face restoration applications are selected to create pseudo HQ labels out of LQ images:

\begin{itemize}
	\item\textbf{Ni-wo-dang-nian}: \url{https://android.myapp.com/myapp/detail.htm?apkName=com.bigwinepot.nwdn\&ADTAG=mobile}
	
	\item\textbf{Remini - Photo Enhancer}:
	\url{https://play.google.com/store/apps/details?id=com.bigwinepot.nwdn.international\&hl=en}
	
	\item\textbf{RetouchMe}: \url{https://retouchme.com/service/photo-restore-app}
	
\end{itemize}




\bibliographystyle{IEEEtran}
\bibliography{ref}

\begin{IEEEbiography}[
	{\includegraphics[width=1in,height=1.25in,clip,keepaspectratio]{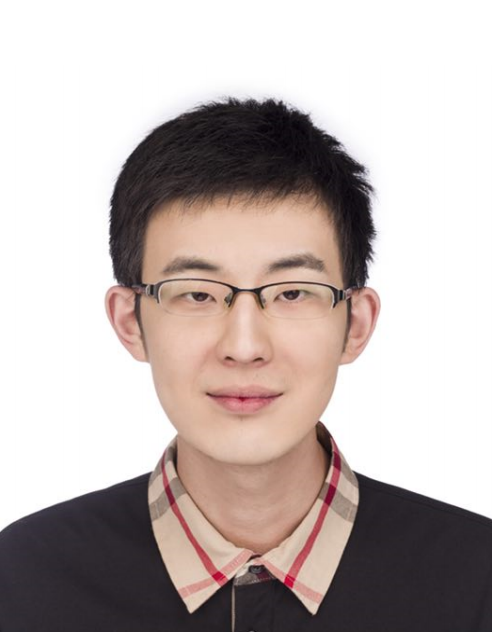}}
]{Lingbo Yang} received the BS degree in mathematics and applied mathematics from Peking University in 2016. He is currently
pursuing the PhD degree at Institute of Digital Media, Peking University. He has been interning at DAMO Academy, Alibaba Group since 2019. His research interests include deep generative models, image restoration and editing, and human pose transfer.
\end{IEEEbiography}

\begin{IEEEbiography}[
	{\includegraphics[width=1in,height=1.25in,clip,keepaspectratio]{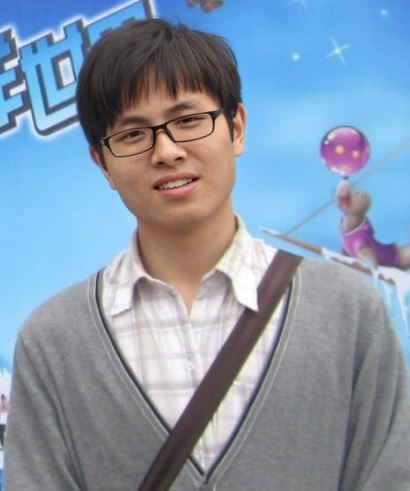}}
	]{Pan Wang} received the B.S. degree in information and control engineering from China University of Petroleum, Qingdao, China, in 2013, and the M.S. degree in computer science with the School of Electronic, Electrical and Communication Engineering, University of Chinese Academy of Sciences, Beijing, China, in 2018. He is now a computer vision algorithm engineer at Alibaba DAMO academy. His current research interests include deep generative models, image restoration, video inpainting and object tracking.
\end{IEEEbiography}

\begin{IEEEbiography}[
	{\includegraphics[width=1in,height=1.25in,clip,keepaspectratio]{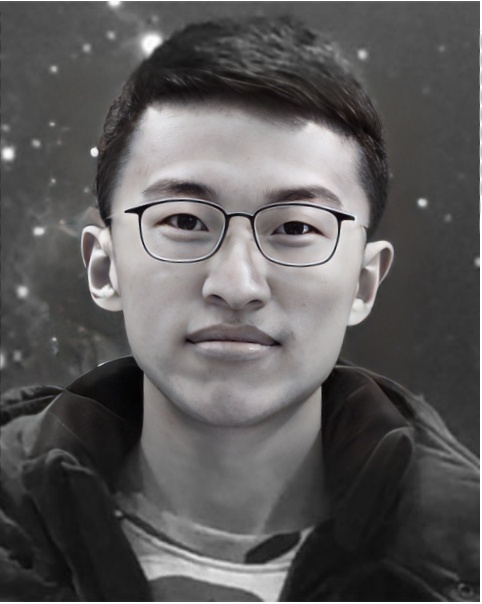}}
	]{Zhanning Gao} received the BS degree in automatic control engineering from Xi’an Jiaotong University, Xi’an, China, in 2012. He is currently
	working toward the PhD degree in the Institute of
	Artificial Intelligence and Robtics, Xi’an Jiaotong
	University. He was a research intern in Visual
	Computing Group, Microsoft Research Asia from
	2015 to 2017. His research interests include compact image/video representation, large scale content based multimedia retrieval, and complex
	event video analysis.
\end{IEEEbiography}

\begin{IEEEbiography}[
	{\includegraphics[width=1in,height=1.25in,clip,keepaspectratio]{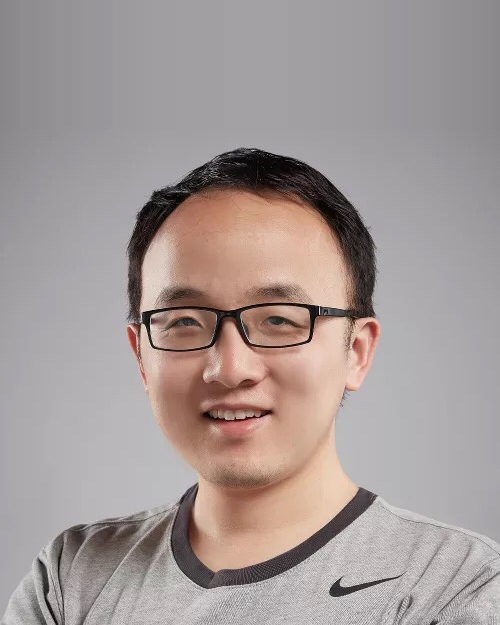}}
]{Peiran Ren} received his BSc and PhD degree from Tsinghua University, China, in 2008 and 2014 respectively. He is now a senior algorithm engineer at Alibaba Damo Acadamy. His research interests include image and video enhancement and processing, computer aided design, real-time rendering, and appearance acquisition.
\end{IEEEbiography}

\begin{IEEEbiography}[
	{\includegraphics[width=1in,height=1.25in,clip,keepaspectratio]{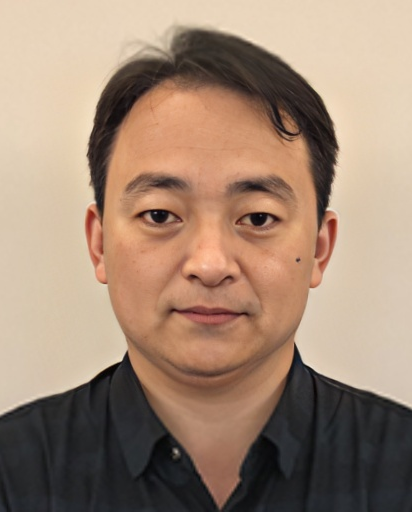}}
	]{Shanshe Wang} received the B.S. degree from the
	Department of Mathematics, Heilongjiang University, Harbin, China, in 2004, the M.S. degree in computer software and theory from Northeast Petroleum
	University, Daqing, China, in 2010, and the Ph.D.
	degree in computer science from the Harbin Institute
	of Technology. He held a post-doctoral position with
	Peking University from 2016 to 2018. He joined
	the School of Electronics Engineering and Computer Science, Institute of Digital Media, Peking
	University, Beijing, where he is currently a Research
	Assistant Professor. His current research interests include video compression
	and image and video quality assessment.
\end{IEEEbiography}
\vfill

\begin{IEEEbiography}[
	{\includegraphics[width=1in,height=1.25in,clip,keepaspectratio]{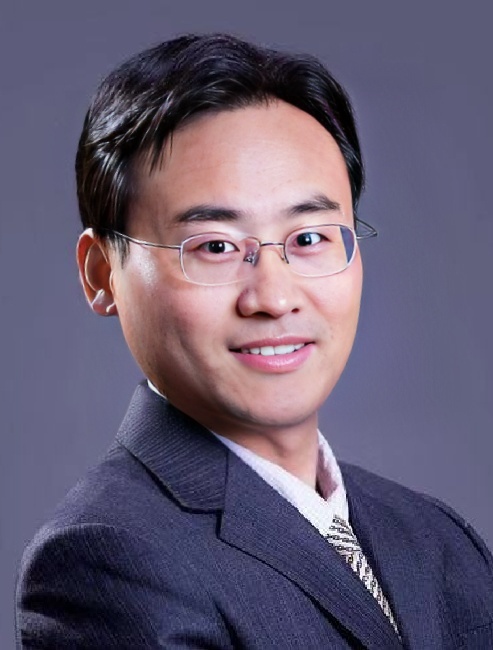}}
	]{Siwei Ma} (Senior Member, IEEE) received the B.S.
	degree from Shandong Normal University, Jinan,
	China, in 1999, and the Ph.D. degree in computer
	science from the Institute of Computing Technology, Chinese Academy of Sciences, Beijing, China,
	in 2005. He held a postdoctoral position with the
	University of Southern California, Los Angeles, CA,
	USA, from 2005 to 2007. He joined the School
	of Electronics Engineering and Computer Science,
	Institute of Digital Media, Peking University, Beijing, where he is currently a Professor. He has
	authored over 200 technical articles in refereed journals and proceedings in
	image and video coding, video processing, video streaming, and transmission.
	He is an Associate Editor of the IEEE TRANSACTIONS ON CIRCUITS AND
	SYSTEMS FOR VIDEO TECHNOLOGY and the Journal of Visual Communication and Image Representation.
\end{IEEEbiography}

\begin{IEEEbiography}[
	{\includegraphics[width=1in,height=1.25in,clip,keepaspectratio]{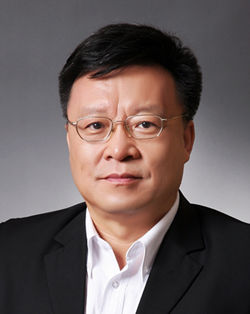}}
	]{Wen Gao} (Fellow, IEEE) received the Ph.D. degree
	in electronic engineering from The University of
	Tokyo, Japan, in 1991. He was a Professor of
	computer science with the Harbin Institute of Technology from 1991 to 1995 and a Professor with the
	Institute of Computing Technology, Chinese Academy of Sciences, from 1996 to 2006. He is currently
	a Professor of computer science with Peking University, Beijing, China. He has authored extensively,
	including five books and more than 600 technical
	articles in refereed journals and conference proceedings in the areas of image processing, video coding and communication,
	pattern recognition, multimedia information retrieval, multimodal interface,
	and bioinformatics. He chaired a number of prestigious international conferences on multimedia and video signal processing, such as IEEE ISCAS,
	ICME, and the ACM Multimedia, and also served on the advisory and
	technical committees for numerous professional organizations. He served
	or serves on the Editorial Board for several journals, such as the IEEE
	TRANSACTIONS ON CIRCUITS AND SYSTEMS FOR VIDEO TECHNOLOGY,
	the IEEE TRANSACTIONS ON MULTIMEDIA, the IEEE TRANSACTIONS ON
	IMAGE PROCESSING. the IEEE TRANSACTIONS ON AUTONOMOUS MENTAL
	DEVELOPMENT, the EURASIP Journal of Image Communications, and the
	Journal of Visual Communication and Image Representation.
\end{IEEEbiography}
\vfill




%

\end{document}